\documentclass{amia}
\usepackage{lipsum} %Remove if not needed

\usepackage{graphicx, wrapfig,lipsum}
\usepackage{amsmath}
\usepackage{algorithm} 
\usepackage{algpseudocode} 
\usepackage{multirow}
\usepackage{booktabs}
\usepackage{amsfonts}
\usepackage{enumitem}
\usepackage[labelfont=bf]{caption}
\begin{document}

\title{DR-VIDAL - Doubly Robust Variational Information-theoretic Deep Adversarial Learning for Counterfactual Prediction and Treatment Effect Estimation on Real World Data}

\author{Shantanu Ghosh MS$^1$, Zheng Feng PhD$^2$, Jiang Bian PhD$^2$, Kevin Butler PhD$^2$, 
Mattia Prosperi PhD$^2$}

\institutes{
    $^1$ University of Pittsburgh, Pittsburgh, PA; $^2$University of Florida, Gainesville, FL
}

\maketitle

\section*{Abstract}

\textit{Determining causal effects of interventions onto outcomes from real-world, observational (non-randomized) data, e.g., treatment repurposing using electronic health records, is challenging due to underlying bias. Causal deep learning has improved over traditional techniques for estimating individualized treatment effects (ITE). We present the Doubly Robust Variational Information-theoretic Deep Adversarial Learning (DR-VIDAL), a novel generative framework that combines two joint models of treatment and outcome, ensuring an unbiased ITE estimation even when one of the two is misspecified. DR-VIDAL integrates: (i) a variational autoencoder (VAE) to factorize confounders into latent variables according to causal assumptions; (ii) an information-theoretic generative adversarial network (Info-GAN) to generate counterfactuals; (iii) a doubly robust block incorporating treatment propensities for outcome predictions. On synthetic and real-world datasets (Infant Health and Development Program, Twin Birth Registry, and National Supported Work Program), DR-VIDAL achieves better performance than other non-generative and generative methods. In conclusion, DR-VIDAL uniquely fuses causal assumptions, VAE, Info-GAN, and doubly robustness into a comprehensive, performant framework. Code is available at:} \url{https://github.com/Shantanu48114860/DR-VIDAL-AMIA-22} \textit{under MIT license.} 

% \textit{Abstract text goes here, justified and in italics.  The abstract would normally be one paragraph long.  See Table~\ref{tab:submission}. for appropriate abstract length by submission type.}

\section*{Introduction}
Understanding causal relationships and evaluating effects of interventions to achieve desired outcomes is key to progress in many fields, especially in medicine and public health. A typical scenario is to determine whether a treatment (e.g., a lipid-lowering medication) is effective to reduce the risk of or cure an illness (e.g., cardiovascular disease). Randomized controlled trials (RCTs) are considered the best practice for evaluating causal effects \cite{Sibbald201}. However, RCTs are not always feasible, due to ethical or operational constraints. For instance, if one wanted to evaluate whether college education is the cause of good salary, it would not be ethical to randomly pick teenagers and randomize their admission to college. So, in many cases, the only usable data sources are observational data, i.e., real-world data collected retrospectively and not randomized. Unfortunately, observational data are often plagued with various biases --since the data generation processes are largely unknown-- such as confounding (i.e., spurious causal effects on outcomes by features that are correlated with a true unmeasured cause) and colliders (i.e., mistakenly including effects of an outcome as predictors), making it difficult to infer causal claims \cite{hernan2019causal}. Another problem is that, in both RCTs and observational datasets, only factual outcomes are available, since clearly an individual cannot be treated and non-treated at the same time. Counterfactuals are alternative predictions that respond to the question \emph{``what outcome would have been observed if a person had been given a different treatment?"}  If models are biased, counterfactual predictions can be wrong, and interventions can be ineffective or harmful \cite{prosperi2020}. In both RCT-based and real-world based studies, two types of treatment effects are usually considered: (i) the average treatment effect (ATE), which is  population-based and represents the difference in average treatment outcomes between the treatment and controls; and (ii) the individualized treatment effect (ITE), which represents the difference in treatment outcomes for a single observational unit with the same background covariates \cite{hoogland21}. When there is suspected heterogeneity, stratified ATEs, or conditional ATEs, can be calculated.
Traditional statistical approaches for estimating treatment effects, taking into account possible bias from pre-treatment characteristics, include propensity score matching (PSM) and inverse probability weighting (IPW) \cite{austin2011introduction}.
The propensity score is a scalar estimate representing the conditional probability of receiving the treatment, given a set of measured pre-treatment covariates. By matching (or weighting) treated and control subjects according to their propensity score, a balance in pre-treatment covariates is induced, mimicking a randomization of the treatment assignment. However, the PSM approach only accounts for measured covariates, and latent bias may remain after matching \cite{garrido2014}. Recently, deep learning-based methods have been used to estimate propensity scores~\cite{ghosh2021propensity, ghosh2021deep}. PSM has been historically implemented with logistic-linear regression, coupled with different feature selection methods in the presence of high-dimensional datasets \cite{tian2018evaluating}. A problem with PSM is that it often decreases the sample size due to matching, while IPW can be affected by skewed, heavy-tailed weight distributions. Machine learning approaches have been introduced more recently, e.g., Bayesian additive regression trees \cite{hill2011bayesian} and counterfactual random forests \cite{wager2018estimation}. Big data also led to the flourishing of causal deep learning \cite{johansson2016learning}. Notable examples include the Treatment-Agnostic Representation Network (TARNet) \cite{pmlr-v70-shalit17a}, Dragonnet \cite{shi2019adapting}, Deep Counterfactual Network with Propensity-Dropout (DCN-PD) \cite{alaa2017deep}, Generative Adversarial Nets for inference of Individualized Treatment Effects (GANITE) \cite{Yoon2018}, Causal Effect Variational Autoencoder (CEVAE) \cite{louizos2017causal}, and Treatment Effect by Disentangled Variational AutoEncoder (TEDVAE) \cite{zhang2020treatment}.

\paragraph{Contribution}

\begin{wrapfigure}{r}{5.5cm}
  \includegraphics[width=0.2\textwidth]{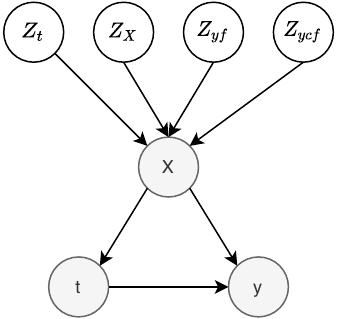}
  \caption{Directed acyclic graph modeling the causal relationships among treatment $t$, outcome $y$ and pre-treatment covariates $X$, under the latent space $Z$.}
  \label{fig: causal_graph}
\end{wrapfigure} 

This work introduces a novel deep learning approach for ITE estimation and counterfactual prediction on real-world observational data, named the \emph{Doubly Robust Variational Information-theoretic Deep Adversarial Learning} (DR-VIDAL). 
Motivated from Makhzani \emph{et al.}  \cite{makhzani2015adversarial}, we use a lower-dimensional neural representation of the input covariates to generate counterfactuals to improve convergence. We assume a causal graph on top of the covariates where the covariates $X$ are generated from 4 independent latent variables $Z_t, Z_{ycf}, Z_{yf}$ and $Z_x$ indicating latents for treatment, counterfactual, factual outcomes and observed covariates respectively, shown in Figure \ref{fig: causal_graph}. In generating the representations, we use a variational autoencoder (VAE) to infer the latent variables from the covariates in unsupervised manner and feed the learned lower-dimensional representation from the VAE to a generative adversarial network (GAN). Also, to counter the loss of the predictive information while generating the counterfactuals, we aim to maximize the mutual information between the learned representations and the output of the generator. We add this as a regularizer to the generator loss to obtain more robust counterfactuals. Finally, we incorporate a doubly robust network head to estimate the ITE, improving in loss convergence. As DR-VIDAL generates the counterfactual outcomes, we minimise the supervised loss for both the factual and the counterfactual outcomes to estimate ITE more accurately.

The main features of DR-VIDAL are, in summary:
\setlist{nolistsep}
\begin{itemize}[noitemsep]
  \item Incorporation of an underlying causal structure where the observed pre-treatment covariate set $X$ is decomposed into four independent latent variables $Z_t, Z_X, Z_{yf}, Z_{ycf}$, inducing confounding on both the treatment and the outcome (Figure \ref{fig: causal_graph}). 
  \item Latent variables are inferred using a VAE \cite{kingma2013auto}.
  \item A GAN \cite{goodfellow2014generative} with variational information maximization \cite{chen2016infogan} generates (synthetic) complete tuples of covariates, treatment, factual and counterfactual outcomes.
  \item Individual treatment effects are estimated on complete datasets with a downstream, four-headed deep learning block which is doubly robust \cite{funk2011doubly, dudik2014doubly}.
\end{itemize}
To our knowledge, this is the first time in which VAE, GAN, information theory and doubly robustness are amalgamated into a counterfactual prediction method. By performing test runs on synthetic and real-world datasets (Infant Health and Development Program, Twin Birth Registry, and National Supported Work Program), we show that DR-VIDAL can outperform a number of state-of-art tools for estimating ITE. DR-VIDAL is implemented in Pytorch and the code is available at: \url{https://github.com/Shantanu48114860/DR-VIDAL-AMIA-22} under MIT license. In the repository, we also provide an online technical supplement (OTS) with full details on the architectural design, derivation of equations, and additional experimental results.

% This template should be used as a starting point for AMIA submissions.  A number of Word styles, all beginning with the word ``AMIA", are available for use in your submissions.

% It is important to review the AMIA Call for Participation where types of submissions considered and general requirements for each submission type are listed. All submissions must conform to the format and presentation requirements described in the CFP and at the submission site.

% \caption{A wrapped figure going nicely inside the text.}\label{wrap-fig:1}
% \includegraphics[width=5.5cm]{sample}

% \begin{wrapfigure}{r}{5.5cm}
% % \begin{figure}
% %   \centering
%   \includegraphics[width=0.2\textwidth]{Causal_Graph.pdf}
%   \caption{Directed acyclic graph modeling the causal relationships among treatment $t$, outcome $y$ and pre-treatment covariates $X$, under the latent space $Z$.}
%   \label{fig: causal_graph}
% % \end{figure}
% \end{wrapfigure} 

\section*{Problem Formulation}
We use the \emph{potential outcomes} framework \cite{rubin1974, rosenbaum1983}. Let us consider a treatment $t$ (binary for ease of reading, but the theory can be extended to multiple treatments) that can be prescribed to a population sample of size $N$. The individuals are characterized by a set of pre-treatment background covariates $\boldsymbol{X}$, and a health outcome $Y$ is measured after treatment. We define each subject $i$ with the tuple \{$\boldsymbol{X},T,Y$\}$_{i=1}^{N}$, where $Y_{i}^0$ and $Y_{i}^1$ are the potential outcomes when applying treatments $T_i=0$ and $T_i=1$, respectively. The ITE $\tau(\textbf{x})$ for subject $i$ with pre-treatment covariates $\boldsymbol{X}_i=\textbf{x}$, is defined as the difference in the average potential outcomes under both treatment interventions (i.e., treated vs. not treated), conditional on $\textbf{x}$, i.e.,
\begin{equation}
    \tau(\textbf{x})=\mathbb{E}[Y_{i}^1 - Y_{i}^0 \mid \boldsymbol{X}_i = \textbf{x}]
\end{equation}
The ITE cannot be calculated directly give the inaccessibility of both potential outcomes, as only factual outcomes can be observed, while the others (counterfactuals) can be considered as missing values. However, when the potential outcomes are made independent of the treatment assignment, conditionally on the pre-treatment covariates, i.e., \{$Y^1,Y^0$\} $\perp T \mid \boldsymbol{X}$, the ITE can then be estimated as
$
\tau(\textbf{x})=\mathbb{E}[Y^1 \mid T=1, \boldsymbol{X} = \textbf{x}] - \mathbb{E}[Y^0 \mid T=0, \boldsymbol{X} = \textbf{x}] 
= \mathbb{E}[Y \mid T=1, \boldsymbol{X} = \textbf{x}] - \mathbb{E}[Y \mid T=0, \boldsymbol{X} = \textbf{x}]
$. 
Such an assumption is called the strongly ignorable treatment assignment (SITA) assumption \cite{imbens2000role, pearl2016causal}. By further averaging over the distribution of $\boldsymbol{X}$, the ATE $\tau_{01}$ can be calculated as
\begin{equation}
\label{ite}
    \tau_{01}=\mathbb{E}[\tau(\boldsymbol{X})]=\mathbb{E}[Y\mid T=1]-\mathbb{E}[Y\mid T=0]
\end{equation}
ITE and ATE can be calculated with stratification matching of $\textbf{x}$ in treatment and control groups, but the calculation becomes unfeasible as the covariate space increases in dimensions. The propensity score $\pi(x)$ represents the probability of receiving the treatment $T=1$ conditioned on the pre-treatment covariates $X=x$, denoted as
$\pi(\textbf{x})$ =  $P(T = 1\mid \boldsymbol{X} = \textbf{x})$ \cite{rosenbaum1983}. The propensity score can be calculated using a regression function, e.g., logistic. ITE/ATE can then be calculated by matching (PSM) or weighting (IPW) instances through $\pi(\textbf{x})$, in a doubly robust way \cite{porter2011relative}, or through myraid approaches \cite{chipman2010bart, wager2018estimation, athey2016recursive, lu2018estimating, porter2011relative, dehejia2002propensity, lunceford2004stratification, crump2008nonparametric}. In the next section, we describe approaches based on deep learning.

\begin{figure*}
\centering
\includegraphics[width=0.7\textwidth]{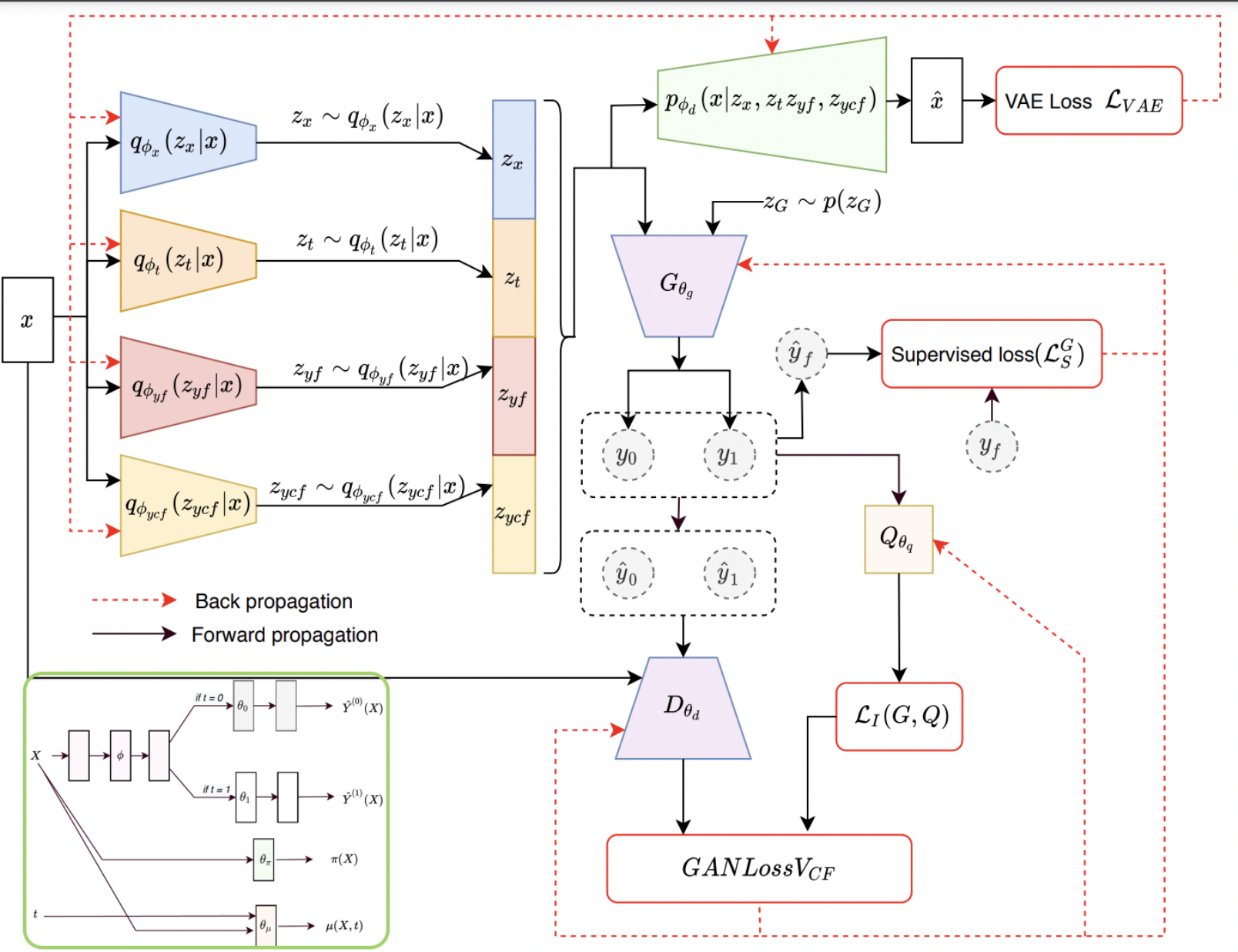} 
\caption{Architecture of DR-VIDAL incorporating the variational autoencoder inferring the latent space (VAE), the generative adversarial network for calculating the counterfactual outcomes (GAN), and the doubly robust module (green box) for estimating ITE.}
\label{fig: vae_gan}
\end{figure*}

\section*{Related Work}
Alaa and Van der Schaar \cite{pmlr-v80-alaa18a} characterized the conditions and the limits of treatment effect estimation using deep learning. The sample size plays an important role, e.g., estimations on small sample sizes are affected by selection bias, while on large sample sizes, they are affected by algorithmic design. Our work builds up on the ITE estimation approaches of CEVAE \cite{louizos2017causal}, DCN-PD \cite{alaa2017deep}, Dragonnet \cite{shi2019adapting}, GANITE \cite{Yoon2018}, TARNet \cite{pmlr-v70-shalit17a}, and TEDVAE \cite{zhang2020treatment}. DCN-PD is a doubly robust, multitask network for counterfactual prediction, where propensity scores are used to determine a dropout probability of samples to regularize training, carried out in alternating phase, using treated and control batches. CEVAE uses VAE to identify latent variables from an observed pre-treatment vector and to generate counterfactuals. TARNet aims to provide an upper bound effect estimation by balancing the distributions of treated and controls --with a weight indemnifying group imbalance-- within a high dimensional covariate space, but it does not exploit counterfactuals, and only minimises the factual loss function. Dragonnet is a modified TARNet with targeted regularization based on propensity scores. GANITE generates proxies of counterfactual outcomes from covariates and random noise using a GAN, and feeds them to an ITE generator. For both GANITE and TARNet, in presence of high-dimensional data, the loss could be hard to converge. TEDVAE \cite{zhang2020treatment} uses a variational autoencoder to infer hidden latent variables from proxies using a causal graph similar to CEVAE. In the next sections, we discuss in detail the novelty of DR-VIDAL and the differences in the architectural design and training mechanisms with respect to the aforementioned approaches.

\section*{Proposed Methodology}
DR-VIDAL architecture can be summarized in three components: (1) a VAE inferring the latent space, (2) a GAN generating the counterfactual outcomes, and (3) a doubly robust module estimating ITE. The architectural layout is schematized in Figure \ref{fig: vae_gan}, while the algorithmic details are given in the OTS.

\paragraph*{Latent variable inference with VAE.}
We assume that the observed covariates \textbf{X} = \textbf{x} with treatment assignment $T = t$ factual and counterfactual outcomes $Y_f = y_f$ and $Y_{cf} = y_{cf}$ respectively, are generated from an independent latent space \textbf{z}, composed by $\textbf{z}_x$ $\sim p(\textbf{z}_x)$, $z_t \sim p(\textbf{z}_t)$, $\textbf{z}_{yf} \sim p(\textbf{z}_{yf})$, and $\textbf{z}_{ycf}\sim p(\textbf{z}_{ycf})$, which denote the latent variables for the covariates \textbf{x}, treatment indicator t, and factual outcomes $y_{f}$ and $y_{cf}$, respectively. This decomposition follows the causal structure shown in Figure \ref{fig: causal_graph}. The goal is to infer the posterior distribution $p(\textbf{z}_x, \textbf{z}_t, \textbf{z}_{yf}, \textbf{z}_{ycf}|\textbf{x})$, which is harder to optimize. We use the theory of variational inference \cite{blei2017variational} to learn the variational posteriors $q_{\phi_x}(\textbf{z}_x|\textbf{x})$, $q_{\phi_t}(\textbf{z}_t|\textbf{x})$, $q_{\phi_{yf}}(\textbf{z}_{yf}|\textbf{x}), q_{\phi_{ycf}}(\textbf{z}_{ycf}|\textbf{x})$, using 4 different neural network encoders with parameters $\phi_x, \phi_t, \phi_{yf},$ and $\phi_{ycf}$, respectively. Using the latent factors sampled from the learned variational posteriors, we reconstruct \textbf{x} by estimating the likelihood $p_{\phi_d}(\textbf{x}|\textbf{z}_x, \textbf{z}_t, \textbf{z}_{yf}, \textbf{z}_{ycf})$ via a single decoder parameterized by $\phi_d$. The latent factors, assumed to be Gaussian, are defined as follows:
\begin{align}
p(\textbf{z}_x) &= \prod_{i=1} ^ {D_{z_x}} \mathcal{N}(z_{x_i}|0, 1) ;
  &p(\textbf{z}_t) &= \prod_{i=1} ^ {D_{z_t}} \mathcal{N}(z_{t_i}|0, 1) \\
p(\textbf{z}_{yf}) &= \prod_{i=1} ^ {D_{z_{yf}}} \mathcal{N}(z_{{yf}_i}|0, 1) ;
  &p(\textbf{z}_{ycf}) &= \prod_{i=1} ^ {D_{z_{ycf}}} \mathcal{N}(z_{{ycf}_i}|0, 1)
\end{align}
where $D_{z_x}, D_{z_t}, D_{z_{yf}}, D_{z_{ycf}}$ are the dimensions of the latent factors $\textbf{z}_x, \textbf{z}_t, \textbf{z}_{yf}, \textbf{z}_{ycf}$, respectively. The variational posteriors of the inference of models are defined as:
\begin{align}
  q_{\phi_x}(\textbf{z}_x|\textbf{x}) &= \prod_{i=1} ^ {D_{z_x}} \mathcal{N}(\mu = \hat{\mu}_x, \sigma^2 = \hat{\sigma}_x^2) \\
  q_{\phi_t}(\textbf{z}_t|\textbf{x}) &= \prod_{i=1} ^ {D_{z_t}} \mathcal{N}(\mu = \hat{\mu}_t, \sigma^2 =\hat{\sigma}_t^2) \\
  q_{\phi_{yf}}(\textbf{z}_{yf}|\textbf{x}) &= \prod_{i=1} ^ {D_{z_{yf}}} \mathcal{N}(\mu = \hat{\mu}_{yf}, \sigma^2 = \hat{\sigma}_{yf}^2) \\
  q_{\phi_{ycf}}(\textbf{z}_{ycf}|\textbf{x}) &= \prod_{i=1} ^ {D_{z_{ycf}}} \mathcal{N}(\mu = \hat{\mu}_{ycf}, \sigma^2 = \hat{\sigma}_{ycf}^2)
  \label{equ: posterior}
\end{align}
where $\hat{\mu}_x, \hat{\mu}_t, \hat{\mu}_{yf}, \hat{\mu}_{ycf}$ and $\hat{\sigma}_{x}^2, \hat{\sigma}_t^2$, $\hat{\sigma}_{yf}^2$, $\hat{\sigma}_{ycf}^2$ are the means and variances of the Gaussian distributions parameterized by encoders $E_{\phi_x}, E_{\phi_t}, E_{\phi_{yf}}, E_{\phi_{ycf}}$ with parameters $\phi_x, \phi_t, \phi_{yf}, \phi_{ycf}$ respectively.
The overall evidence lower bound (ELBO) loss of the VAE is expressed as $\mathcal{L}_{ELBO}$ in the following equation,
\begin{align*}
    & \mathcal{L}_{ELBO}(\phi_x, \phi_t, \phi_{yf}, \phi_{ycf}; \textbf{x}, \textbf{z}_x, \textbf{z}_t, \textbf{z}_{yf}, \textbf{z}_{ycf}) 
    = \mathbb{E}_{q_{\phi_x}, q_{\phi_t}, q_{\phi_{yf}}, q_{\phi_{ycf}}}[\log p_{\phi_d}(\textbf{x}|\textbf{z}_x, \textbf{z}_t, \textbf{z}_{yf}, \textbf{z}_{ycf})]\\
    & -KL\big(q_{\phi_x}(\textbf{z}_x|\textbf{x}) || p_{\phi_d}(\textbf{z}_x))\big) - KL\big(q_{\phi_t}(\textbf{z}_t|\textbf{x}) || p_{\phi_d}(\textbf{z}_t))\big) \\
    & - KL\big(q_{\phi_{yf}}(\textbf{z}_{yf}|\textbf{x}) || p_{\phi_d}(\textbf{z}_{yf}))\big)  - KL\big(q_{\phi_{ycf}}(\textbf{z}_{ycf}|\textbf{x}) || p_{\phi_d}(\textbf{z}_{ycf}))\big)
\end{align*}
where KL denotes the Kullback–Leibler divergence of two probability distributions. We minimize the optimization function of the VAE as $\mathcal{L}_{VAE}$ to obtain the optimal parameter of the encoders $\phi_x, \phi_t, \phi_{yf}, \phi_{ycf}$, and of the decoder $\phi_d$ as $\mathcal{L}_{VAE}(\phi_x, \phi_t, \phi_{yf}, \phi_{ycf}; \textbf{x}, \textbf{z}_x, \textbf{z}_t, \textbf{z}_{yf}, \textbf{z}_{ycf})$ = $-\mathcal{L}_{ELBO}(\phi_x, \phi_t, \phi_{yf}, \phi_{ycf}; \textbf{x}, \textbf{z}_x, \textbf{z}_t, \textbf{z}_{yf}, \textbf{z}_{ycf})$.

\subparagraph{Generation of counterfactuals via GAN.}
After learning the hidden latent codes $\textbf{z}_x, \textbf{z}_t, \textbf{z}_{yf}, \textbf{z}_{ycf}$ from the VAE, we concatenate the latent codes to form $z_c$, passed to the generator of the GAN block $G_{\theta_g}$, along with a random noise $\textbf{z}_G \sim \mathcal{N}(0, Id)$. $G_{\theta_g}$ is parameterized by $\theta_g$, and it outputs the vector $\overline{y}$ of the potential (factual and counterfactual) outcomes. We replace the factual outcome $y_f$ in the generated outcome vector $\overline{y}$ to form $\hat{y}_0$ and $\hat{y}_1$, which are passed to the counterfactual discriminator $D_{\theta_d}$, along with the true covariate vector \textbf{x}.  $D_{\theta_d}$ is parameterized by $\theta_d$, and is responsible to predict the treatment variable, similarly to GANITE. The loss of the GAN block is defined as:
\begin{align*}
    & V_{GAN} (G, D) = \mathbb{E}_{\textbf{x}, \textbf{z}_G, \textbf{z}_c}\big[t^T\log D(\textbf{x}, G(\textbf{z}_G, \textbf{z}_c)) 
    + (1 - t)^T \log(1 - D(\textbf{x}, G(\textbf{z}_G, \textbf{z}_c)) \big]
    \label{equ:V_CF}
\end{align*}
where $\textbf{x} \sim p(\textbf{x}), \textbf{z}_G \sim p(\textbf{z}_G)$ and $\textbf{z}_c$ denote the concatenated latent codes 
$\textbf{z}_x \sim q_{\phi_x}(\textbf{z}_x|\textbf{x})$, $\textbf{z}_t \sim q_{\phi_t}(\textbf{z}_t|\textbf{x})$, $\textbf{z}_{yf} \sim q_{\phi_{yf}}(\textbf{z}_{yf}|\textbf{x})$ and 
$\textbf{z}_{ycf} \sim q_{\phi_{ycf}}(\textbf{z}_{ycf}|\textbf{x})$.
From $\overline{y}$, we also calculate the predicted factual outcome $\hat{y}_f$. As also done in GANITE, we make sure to include the supervised loss $\mathcal{L}_S^G(y_f, \hat{y}_f)$, which enforces the predicted factual outcome $\hat{y}_f$ to be as close as to the true factual outcome $y_f$.
\begin{align}
    \mathcal{L}_S^G(y_f, \hat{y}_f) = \frac{1}{n} \sum_{i = 1}^n \big(y_f(i) - \hat{y}{_f(i)}\big)^2 
\end{align}
The complete loss function of counterfactual GAN is given by $V_{CF} (G, D) = V_{GAN} (G, D) + \gamma \mathcal{L}_S^G(y_f, \hat{y}_f)$.

We also employ an additional regularization $\lambda I(\textbf{z}_c; G(\textbf{z}_G, \textbf{z}_c))$ to maximize the mutual information between the learned concatenated latent code $\textbf{z}_c$ and the generated output by the generator $ G(\textbf{z}_G, \textbf{z}_c)$, as in \cite{chen2016infogan}.We thus propose to solve the following minimax game:
\begin{equation}
    \min_G \max_D V_{CF\_I} (G, D) = V_{CF}(G, D) + \lambda I(\textbf{z}_c; G(\textbf{z}_G, \textbf{z}_c))
\end{equation}

$I(\textbf{z}_c; G(\textbf{z}_G, \textbf{z}_c))$ is harder to solve because of the presence of the posterior $p(\textbf{z}_c|\textbf{x})$ \cite{chen2016infogan}, so we obtain the lower bound of it using an auxiliary distribution $Q(\textbf{z}_c|\textbf{x})$ to approximate $p(\textbf{z}_c|\textbf{x})$. 

Finally, the optimization function of the counterfactual information-theoretic GAN --\emph{InfoGAN}-- incorporating the variational regularization of mutual information and hyperparameter $\lambda$ is given by:
\begin{equation}
    \min_{G, Q} \max_{D} V_{CF\_infoGAN}(G, D, Q) = V_{CF}(G, D) - \lambda \mathcal{L}_I(G, Q)
    \label{equ: Info_GAN}
\end{equation}

The counterfactual InfoGAN is used to generate the missing counterfactual outcome $y_{cf}$ to form the quadruple \{$\textbf{x}$, $t$, $y_f$, $y_{cf}$\}$_{i=1}^N$ and sent to the doubly robust block to estimate the ITE.

\paragraph*{Information-theoretic GAN optimization.}
The GAN generator $G_{\theta_g}$ works to fool the discriminator $D_{\theta_d}$. To get the optimal Discriminator $D_{\theta_d}^*$, we maximize $V_{CF\_infoGAN}$
\begin{equation}
    \max_D \mathcal{L}^D(\theta_d) =  V_{CF\_infoGAN}(G, D, Q)
    \label{equ: Info_GAN_D}
\end{equation}

To get the optimal generator $G_{\theta_g}^*$, we maximize $V_{CF\_infoGAN}$
\begin{equation}
    \min_{G, Q} \mathcal{L}^G(\theta_g) =  V_{CF\_infoGAN}(G, D, Q)
    \label{equ: Info_GAN_G}
\end{equation}

\paragraph*{Doubly robust ITE estimation.}

As introduced above, the propensity score $\pi(\textbf{x})$ represents the probability of receiving a treatment $T=1$ (over the alternative $T=0$) conditioned on the pre-treatment covariates $X = x$. By combining IPW through $\pi(\textbf{x})$ with outcome regression by both treatment variable and the covariates, Jonsson defined the doubly robust estimation of causal effect \cite{funk2011doubly} as follows:
\begin{align}
    &\hat{\delta}_{DR} = \frac{1}{n}\sum_{i=1}^n\bigg[\frac{y_it_i - (t_i - \pi(x_i))\mu(x_i, t_i)}{\pi(x_i)} -\frac{y_i(1 - t_i) - (t_i - \pi(x_i))\mu(x_i, t_i)}{1-\pi(x_i)}\bigg] 
\end{align}
where $\mu(x, t) = \hat{\alpha_0} + \hat{\alpha_1}x_1 + \hat{\alpha_2}x_2 + \dots + \hat{\alpha_n}x_n + \hat{\delta}t$, and 
    $(t_i - \pi(x_i))\mu(x_i, t_i)$ is used for the IPW estimator.

After getting the counterfactual outcome $y_{cf}$ from the counterfactual GAN to form the quadruple \{$\textbf{x}$, $t$, $y_f$, $y_{cf}$\}$_{i=1}^N$, we pass this as the input to the doubly robust multitask network to estimate the ITE, using the architecture shown in Figure \ref{fig: vae_gan} (green box). To predict the outcomes $y^{(0)}$ and $y^{(1)}$, we use a configuration similar to TARNet,  which contains a number of shared layers, denoted by $f_\phi$, parameterized by $\phi$, and two outcome-specific heads $f_{\theta_0}$ and $f_{\theta_1}$, parameterized by $\theta_0$ and $\theta_1$.

To ensure doubly robustness, we introduce two more heads that predict the propensity score $\pi(\textbf{x}) = \mathbb{P}(T=1|\textbf{x})$ and the regressor $\mu(\textbf{x}, t)$. These two are calculated using two neural networks, parameterized by $\theta_\pi$ and $\theta_\mu$ respectively. 
The factual and counterfactual outcome $y^{(0)}_i$ and $y^{(1)}_i$ of the $i^{th}$ sample are then calculated as:
\begin{align}
    % \hat{y}^{(0)}_i &= f_{\theta_0}(f_{\phi}(\textbf{x}_i)) && \text{if $t_i$ = 0}\\
    % \hat{y}^{(1)}_i &= f_{\theta_1}(f_{\phi}(\textbf{x}_i)) && \text{if $t_i$ = 1}\\
    \hat{y}_f^{(i)} &= t_i(f_{\theta_1}(f_{\phi}(\textbf{x}_i))) + (1-t_i)(f_{\theta_0}(f_{\phi}(\textbf{x}_i))) \\
    \hat{y}_{cf}^{(i)} &= (1-t_i)(f_{\theta_1}(f_{\phi}(\textbf{x}_i))) + t_i(f_{\theta_0}(f_{\phi}(\textbf{x}_i)))
    \label{equ: pred_y}
\end{align}
Next, the predicted loss will be  
% $\mathcal{L}^p_i(\theta_1, \theta_0, \phi)=(\hat{y}_f^{(i)} - y_f^{(i)})^2 + (\hat{y}_{cf}^{(i)} - y_{cf}^{(i)})^2 + \alpha \cdot \text{BinaryCrossEntropy}(\pi(\textbf{x}_i), t_i)$,
\begin{align}
    \mathcal{L}^p_i(\theta_1, \theta_0, \phi) &= (\hat{y}_f^{(i)} - y_f^{(i)})^2 + (\hat{y}_{cf}^{(i)} - y_{cf}^{(i)})^2 \nonumber
    + \alpha \text{BinaryCrossEntropy}(\pi(x_i), t_i)
    \label{equ: pred_loss}
\end{align}
 where $\alpha$ is a hyperparameter. With the help of the propensity score $\pi(\textbf{x})$ and the regressor $\mu(\textbf{x}, T)$, the doubly robust outcomes are calculated as 
\begin{align}
     &\hat{y}_{f_{DR}}^{(i)} = t_i\bigg[ \frac{t_i\hat{y}^{(1)}_i - (t_i - \pi(\textbf{x}_i)\mu(\textbf{x}_i, t_i))}{\pi(\textbf{x}_i)}\bigg] 
      + (1-t_i)\bigg[ \frac{(1-t_i)\hat{y}^{(0)}_i - (t_i - \pi(\textbf{x}_i)\mu(\textbf{x}_i, t_i))}{1- \pi(\textbf{x}_i)}\bigg] \\
     &\hat{y}_{{cf}_{DR}}^{(i)} = (1-t_i)\bigg[ \frac{(1-t_i)\hat{y}^{(1)}_i - (t_i - \pi(\textbf{x}_i)\mu(\textbf{x}_i, t_i))}{\pi(\textbf{x}_i)}\bigg] + t_i\bigg[ \frac{t_i\hat{y}^{(0)}_i - (t_i - \pi(\textbf{x}_i)\mu(\textbf{x}_i, t_i))}{1- \pi(\textbf{x}_i)}\bigg]
    \label{equ: outcome_DR}
\end{align}

The doubly robust loss $\mathcal{L}_i^{DR} (\theta_1, \theta_0, \theta_\phi, \theta_\mu, \phi)$ is calculated as:
\begin{align}
    \mathcal{L}_i^{DR} (\theta_1, \theta_0, \theta_\pi, \theta_\mu, \phi) = (\hat{y}_{f_{DR}}^{(i)} - y_f^{(i)})^2 + 
    (\hat{y}_{{cf}_{DR}}^{(i)} - y_{cf}^{(i)})^2
    \label{equ: loss_DR}
\end{align}

Finally, the loss function of the ITE is:
\begin{align}
    &\mathcal{L}^{ITE}(\theta_1, \theta_0, \theta_\pi, \theta_\mu ,\phi) = \frac{1}{n}\sum_{i=1}^n \bigg(\mathcal{L}^p_i+ \beta\mathcal{L}_i^{DR}\bigg)
    \label{eq: loss_Final_DR}
\end{align}
where $\beta$ is a hyperparameter and the whole network is trained using end-to-end strategy.

\section*{Experimental Setup}

\paragraph{Synthetic datasets.} We conduct performance tests on two synthetic data experiments.
The first uses the same data generation process devised for CEVAE \cite{louizos2017causal}. We generate a marginal distribution \textbf{x} as a mixture of Gaussians from the 5-dimensional latent variable \textbf{z}, indicating each mixture component. The details of the synthetic dataset using this process is discussed in the OTS. Datasets of sample size \{1000, 3000, 5000, 10000, 30000\} are generated, and divided into 80-20 \% train-test split.
% \begin{align}
% \textbf{z}_i \sim Bern(0.5); & \qquad
% \textbf{x}_i|\textbf{z}_i \sim \mathcal{N}(\textbf{z}_i, \sigma^2_{5}\textbf{z}_i + \sigma^2_{3}(1 - \textbf{z}_i)) \nonumber\\
% t_i|\textbf{z}_i &\sim Bern(0.75\textbf{z}_i + 0.25(1 - \textbf{z}_i)) \nonumber\\
% \textbf{y}_i|_i, \textbf{z}_i &\sim Bern(Sigmoid(3(\textbf{z}_i + 2(2t_i - 1))))
% \label{equ: CEVAE_data}
% \end{align}
In the second experimental setting, we amalgamate the synthetic data generation process by CEVAE with that of GANITE \cite{Yoon2018}, to model the more complex causal structure illustrated in Figure \ref{fig: causal_graph}. We sample 7-, 1-, 1-, and 1-dimensional vectors for $\textbf{z}_x$, $\textbf{z}_t$, $\textbf{z}_{yf}$, and $\textbf{z}_{ycf}$ from Bernoulli distributions, and then collate them into $x$. From the covariates $x$, we simulate the treatment assignment $t$ and the potential outcomes $y$ as described in the GANITE paper. We generate multiple synthetic datasets for sample sizes \{1000, 3000, 5000, 10000, 30000\}, also divided into 80-20 \% splits. Equations for both data generating processes are provided in the OTS.

% \begin{align}
% \textbf{z}_x \sim Bern(0.5); & \qquad \textbf{z}_t \sim Bern(0.5) \nonumber\\
% \textbf{z}_{yf} \sim Bern(0.5); & \qquad \textbf{z}_{ycf} \sim Bern(0.5) \nonumber\\
% \textbf{x}_x|\textbf{z}_x &\sim \mathcal{N}(\textbf{z}_x, 5(\textbf{z}_x) + 3(1 - \textbf{z}_x)) \nonumber\\
% \textbf{x}_t|\textbf{z}_t &\sim \mathcal{N}(\textbf{z}_x, 2(\textbf{z}z_t) + 0.5(1 - \textbf{z}_t)) \nonumber\\
% \textbf{x}_{yf}|\textbf{z}_{yf} &\sim \mathcal{N}(\textbf{z}_{yf}, 10(\textbf{z}_{yf}) + 6(1 - \textbf{z}_{yf})) \nonumber\\
% \textbf{x}_{ycf}|\textbf{z}_{ycf} &\sim \mathcal{N}(\textbf{z}_{ycf}, 10(\textbf{z}_{ycf}) + 6(1 - \textbf{z}_{ycf})) \nonumber\\
% \textbf{w}_t^T \sim \mathcal{U}((-0.1, 0.1)^{10 \text{x} 1}); & \qquad
% \textbf{n}_t \sim \mathcal{N}(0, 0.1)\nonumber\\
% \textbf{w}_y^T \sim \mathcal{U}((-1, 1)^{10 \text{x} 2}); & \qquad 
% \textbf{n}_y \sim \mathcal{N}(0^{2 \text{x} 1}, 0.1 \text{x} \mathcal{I}^{2 \text{x} 2})\nonumber\\
% t|x \sim Bern(Sigmoid(\textbf{w}_t^T\textbf{x} + \textbf{n}_t)); & \qquad
% \textbf{y}|x \sim \textbf{w}_y^T\textbf{x} + \textbf{n}_y
% \label{equ: DR_VIDAL_data}
% \end{align}

\paragraph{Real-world datasets.} 
We use three popular real-world benchmark datasets: the Infant Health and Development Program (IHDP) dataset \cite{hill2011bayesian}, the Twins dataset \cite{almond2005costs}, and the Jobs dataset \cite{lalonde1986evaluating}. The IHDP and Twins two are semi-synthetic, and simulated counterfactuals to the real factual data are available. These datasets have been also designed and collated to meet specific treatment overlap condition, nonparallel treatment assignment, and nonlinear outcome surfaces \cite{hill2011bayesian, pmlr-v70-shalit17a, louizos2017causal, Yoon2018}. In detail, IHDP collates data from a multi-site RCT evaluating early intervention in premature, low birth infants, to decrease unfavorable health outcomes. The dataset is composed by 110 treated subjects and 487 controls, with 25 covariates. The Twins dataset is based on records of twin births in the USA from 1989-1991, where the outcome is mortality in the first year, and treatment is heavier weight, comprising 4553 treated, 4567 controls, with 30 covariates. The Jobs study (1978-1978) investigates if a job training program intervention affects earnings after a two-year period, and comprises 237 treated, 2333 controls, with 17 covariates. For all the real-world datasets, we use the same experimental settings described in GANITE, where the datasets are divided into 56/24/20 \% train-validation-test splits. We run 1000, 10 and 100 realizations of IHDP, Jobs and Twins datasets, respectively. 

\begin{figure}[t]
\centering
\includegraphics[width=0.8\textwidth]{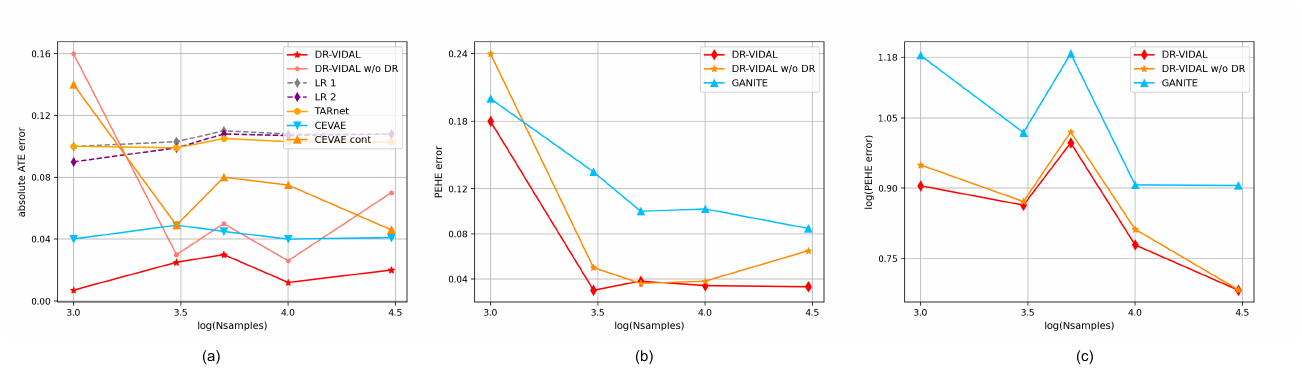} 
\caption{Panel (a): performance (ATE) of DR-VIDAL vs. all other models on samples from the generative process of CEVAE. Panel (b) and (c): performance (PEHE) of DR-VIDAL with or without the doubly robust (DR, w/o DR) block vs. GANITE on  samples from the generative process of CEVAE-GANITE.}
\label{fig: Graphs}
\end{figure}

\paragraph{Model fit and test details.} Consistent with prior studies \cite{hill2011bayesian, pmlr-v70-shalit17a, Yoon2018}, we report the error on the ATE $\epsilon_{ATE}$, and the expected Precision in Estimation of Heterogeneous Effect (PEHE), $\epsilon_{PEHE}$, for IHDP and Twins datasets, since factual and the counterfactual outcomes are available. For the Jobs dataset, as the counterfactual outcome does not exist, we report the policy risk $R_{pol}(\pi)$, and the error on the average treatment effect on the treated (ATT) $\epsilon_{ATT}$, as indicated in \cite{pmlr-v70-shalit17a, Yoon2018}. The training details and the hyperparameters of the individual networks are given in the OTS. We compared DR-VIDAL with TARNet, CEVAE, and GANITE. In addition, for real-world datasets, we compare: least squares regression with treatment as a covariate (OLS/LR1); separate least squares regression for each treatment (OLS/LR2); balancing linear regression (BLR) \cite{johansson2016learning}; k-nearest neighbor (k-NN)
\cite{crump2008nonparametric}; Bayesian additive regression trees (BART) \cite{chipman2010bart}; random and causal forest (R Forest, C Forest) \cite{wager2018estimation}; balancing
neural network (BNN) \cite{johansson2016learning}; counterfactual regression with Wasserstein distance (CFR$_{WASS}$) \cite{pmlr-v70-shalit17a}.

\section*{Results}
\begin{table*}[h]
\fontsize{7.6pt}{0.35cm}\selectfont
% \small
\centering
\begin{tabular}{p{1.5cm}|l r|l r|l r}
\toprule 
    % & \multicolumn{2}{c}{\textbf{IHDP}} & \multicolumn{2}{|c}{\textbf{Jobs}} \\
      & \multicolumn{2}{c}{\textbf{IHDP($\boldsymbol{\sqrt{\epsilon_{PEHE}}}$)}} & \multicolumn{2}{|c}{\textbf{Twins($\boldsymbol{\sqrt{\epsilon_{PEHE}}}$)}} 
      & \multicolumn{2}{|c}{\textbf{Jobs($\boldsymbol{R_{Pol}}$)}}\\
      & Out-Sample & In-Sample 
      & Out-Sample & In-Sample
      & Out-Sample & In-Sample\\
\midrule 
      OLS/LR1   & 5.8 $\pm$ 0.3* & 5.8 $\pm$ 0.3*  & 0.318 $\pm$ 0.007 & 0.319 $\pm$ 0.005*
      & 0.23 $\pm$ 0.02* & 0.22 $\pm$ 0.00*\\
      OLS/LR2   & 2.5 $\pm$ 0.1* & 2.4 $\pm$ 0.1*  & 0.320 $\pm$ 0.003* & 0.320 $\pm$ 0.001*
      & 0.24 $\pm$ 0.01* & 0.21 $\pm$ 0.00*\\
      BLR       & 5.8 $\pm$ 0.3* & 5.8 $\pm$ 0.3*  & 0.323 $\pm$ 0.018* & 0.312 $\pm$ 0.002*
      & 0.25 $\pm$ 0.02* & 0.22 $\pm$ 0.01*\\
      k-NN      & 4.1 $\pm$ 0.2* & 2.1 $\pm$ 0.1*  & 0.345 $\pm$ 0.007* & 0.333 $\pm$ 0.003*
      & 0.26 $\pm$ 0.02* & 0.02 $\pm$ 0.00*\\
\midrule 
      BART      & 2.3 $\pm$ 0.1* & 2.1 $\pm$ 0.2* & 0.338 $\pm$ 0.016* & 0.347 $\pm$ 0.009*
      & 0.25 $\pm$ 0.00* & 0.23 $\pm$ 0.02*\\
      R Forest  & 6.6 $\pm$ 0.3* & 4.2 $\pm$ 0.2* & 0.321 $\pm$ 0.005* & 0.306 $\pm$ 0.002
      & 0.28 $\pm$ 0.02* & 0.23 $\pm$ 0.01*\\
      C Forest  & 3.8 $\pm$ 0.2* & 3.8 $\pm$ 0.2* & 0.316 $\pm$ 0.011 & 0.366 $\pm$ 0.003*
      & 0.20 $\pm$ 0.02* & 0.19 $\pm$ 0.00*\\
\midrule 
      BNN      & 2.1 $\pm$ 0.1* & 2.2 $\pm$ 0.1* & 0.321 $\pm$ 0.018* & 0.325 $\pm$ 0.003*
      & 0.24 $\pm$ 0.02* & 0.20 $\pm$ 0.01*\\
      TARNET (TensorFlow)   & 0.95 $\pm$ 0.02* & 0.88 $\pm$ 0.02* & 0.315 $\pm$ 0.003 & 0.317 $\pm$ 0.007
      & 0.21 $\pm$ 0.01* & 0.17 $\pm$ 0.01*\\
      TARNeT (Pytorch)   & 1.10 $\pm$ 0.02* & - & - & -
      & 0.29 $\pm$ 0.06* & -\\
      CFR$_{WASS}$ & 0.76 $\pm$ 0.0* & 0.71 $\pm$ 0.0* & 0.313 $\pm$ 0.008 & 0.315 $\pm$ 0.007
      & 0.21 $\pm$ 0.01* & 0.17 $\pm$ 0.01*\\
\midrule 
      GANITE   & 2.4 $\pm$ 0.4* & 1.9 $\pm$ 0.4* & 0.297 $\pm$ 0.05 & 0.289 $\pm$ 0.005
      & 0.14 $\pm$ 0.01* & 0.13 $\pm$ 0.01*\\
      CEVAE    & 2.6 $\pm$ 0.1* & 2.7 $\pm$ 0.1* & n.r & n.r
      & 0.26 $\pm$ 0.0* & 0.15 $\pm$ 0.0*\\
\midrule
      \textbf{DR-VIDAL}  & \textbf{0.69 $\pm$ 0.06} & \textbf{0.69 $\pm$ 0.05} 
      & \textbf{0.318 $\pm$ 0.008} & \textbf{0.317 $\pm$ 0.002} & \textbf{0.10 $\pm$ 0.01}
      & \textbf{0.09 $\pm$ 0.005} \\
\bottomrule
\end{tabular}
\caption{Performance of $\sqrt{\epsilon_{PEHE}}$ and $R_{Pol}$ (mean $\pm$ st.dev) of
various models (prior tools and DR-VIRDAL) on the IHDP, Twins and Jobs datasets. TARNet was originally developed in TensorFlow. We re-implemented TARNet in Pytorch for IHDP and Jobs dataset. (*) is used to indicate methods that DR-VIDAL shows a statistically significant improvement over}\smallskip
\label{tab: IHDP_Jobs_Twins}
\end{table*}

\paragraph{Synthetic datasets.}
Figure \ref{fig: Graphs} (a), (b) and (c) shows ATE/PEHE results of DR-VIDAL vs. all other models according to the two synthetic data generation processes. In the generative process of CEVAE, the doubly robust version of DR-VIDAL demonstrates lower ATE error than all other models at all sample sizes. When comparing PEHE, DR-VIDAL (both with and without the doubly robust feature) largely outperforms GANITE. In the second synthetic dataset, generated under the more complex assumptions, DR-VIDAL (both with and without the doubly robust feature) outperforms GANITE in terms of PEHE. It is worth noting the potential of DR-VIDAL to better infer hidden representations in comparison to GANITE irrespective of the presence of the doubly robust module.

\paragraph{Real world datasets.}
 In all three IHDP, Jobs and Twins datasets, across all realizations, the information-theoretic, doubly robust configuration of DR-VIDAL yields the best results against all other configurations --with/without information-theoretic optimization and with/without doubly robust loss. The doubly robust loss seems to be responsible for most of the improvement. The absolute gain is small, in the order of 1\%, but the relative gain with respect to the non-doubly robust setup is significant, where the doubly robust module always outperforms its non-doubly robust version, from 55-60\% in IHDP to over 80\% in Twins and Jobs datasets (Figure \ref{fig: boxplots_yf}). Table \ref{tab: IHDP_Jobs_Twins} shows the comparison for the $\sqrt{\epsilon_{PEHE}}$ and $R_{Pol}$ values with the state-of-the-art methods on the three datasets. DR-VIDAL outperforms the other methods on all datasets. On the IHDP and Jobs dataset, DR-VIDAL is the best over all by a larger margin.  Instead, performance increment in the Twins dataset is mild. Even if DR-VIDAL has a large number of parameters, the deconfounding of hidden factors and the adversarial training make it appropriate for datasets with relatively small sample size like IHDP. It is worth noting that DR-VIDAL converges much faster than CEVAE and GANITE, possibly due to the doubly robustness. 
 
\begin{figure*}[t]
\centering
\includegraphics[width=0.75\textwidth]{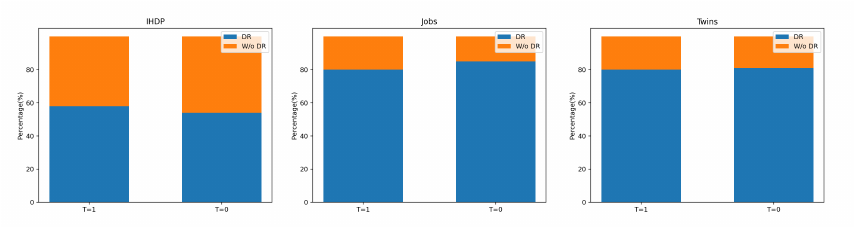} 
\caption{Performance comparison of doubly robust vs. non-doubly robust version of DR-VIDAL. The bar plots show how many times one model setup is better than the other in terms of error on the factual outcome ($y_f$). Panels, from left to right, show results on IHDP, Jobs and Twins datasets (100, 10, 100 iterations), respectively.}
\label{fig: boxplots_yf}
\end{figure*}

%   The t-distributed stochastic neighbor embedding (t-SNE) of representations learned by the VAE of the adversarial module of DR-VIDAL for Twins and Jobs datasets --before and after training-- are shown in Figure \ref{fig: Twins_Jobs_latent}. For all datasets, the t-SNE shows reorganization and cluster tightness (i.e., the data reside on a smaller space) on the treatment, factual and counterfactual outcomes spaces. 

\section*{Conclusions}
DR-VIDAL is a new deep learning approach to causal effect estimation and counterfactual prediction that combines adversarial representation learning, information-theoretic optimization, and doubly robust regression. On the benchmark datasets, both the doubly robust property and information-theoretic optimization of DR-VIDAL improve performance over a basic adversarial setup.

The work has some limitations. First, the causal graph, even if more elaborate than CEVAE, could be improved. For instance, by connecting the $Z$ to $X$ and only to their respective $t$, factual and counterfactual outcome nodes would imply two adjustments set. Another option could be to use the TEDVAE structure in conjunction with out doubly-robust setup. Also, the encoded representation in the VAE does not employ any attention mechanism to identify the most important covariates for the propensity scores, especially with of high-dimensional datasets. Finally, one thing that would be worth evaluating is how Dragonnet would perform as a downstream module of DR-VIDAL, substituting it to our current four-head doubly-robust block.

In conclusion, DR-VIDAL framework is a comprehensive approach to predicting counterfactuals and estimating ITE, and its flexibility (modifiable causal structure and modularity) allows for further expansion and improvement.

\section*{Acknowledgments} This work was in part funded by NIH awards R21CA245858, R01CA246418, R56AG069880, R01AG076234, R01AI145552, R01AI141810, and NSF 2028221.

\makeatletter
\renewcommand{\@biblabel}[1]{\hfill #1.}
\makeatother

% unstr is used to keep citation order
\bibliographystyle{vancouver}
\bibliography{amia}  

\begin{thebibliography}{10}

\bibitem{Sibbald201}
Sibbald B, Roland M.
\newblock Understanding controlled trials: Why are randomised controlled trials
  important?
\newblock BMJ. 1998;316(7126):201.

\bibitem{hernan2019causal}
Hern{\'a}n MA, Robins JM. Causal inference: what if. Boca Raton: Chapman \&
  Hall/CRC; 2020.

\bibitem{prosperi2020}
Prosperi M, Guo Y, Sperrin M, Koopman JS, Min JS, He X, et~al.
\newblock Causal inference and counterfactual prediction in machine learning
  for actionable healthcare.
\newblock Nature Machine Intelligence. 2020;2:369-75.

\bibitem{hoogland21}
Hoogland J, IntHout J, Belias M, Rovers MM, Riley RD, E~Harrell~Jr F, et~al.
\newblock A tutorial on individualized treatment effect prediction from
  randomized trials with a binary endpoint.
\newblock Statistics in Medicine. 2021;40(26):5961-81.
\newblock Available from:
  \url{https://onlinelibrary.wiley.com/doi/abs/10.1002/sim.9154}.

\bibitem{austin2011introduction}
Austin PC.
\newblock An introduction to propensity score methods for reducing the effects
  of confounding in observational studies.
\newblock Multivariate behavioral research. 2011;46(3):399-424.

\bibitem{garrido2014}
Garrido M, et~al.
\newblock Methods for Constructing and Assessing Propensity Scores.
\newblock Health Services Research. 2014;49(5):1701–20.

\bibitem{ghosh2021propensity}
Ghosh S, Boucher C, Bian J, Prosperi M.
\newblock Propensity score synthetic augmentation matching using generative
  adversarial networks (PSSAM-GAN).
\newblock Computer methods and programs in biomedicine update. 2021;1:100020.

\bibitem{ghosh2021deep}
Ghosh S, Bian J, Guo Y, Prosperi M.
\newblock Deep propensity network using a sparse autoencoder for estimation of
  treatment effects.
\newblock Journal of the American Medical Informatics Association.
  2021;28(6):1197-206.

\bibitem{tian2018evaluating}
Tian Y, Schuemie MJ, Suchard MA.
\newblock Evaluating large-scale propensity score performance through
  real-world and synthetic data experiments.
\newblock International journal of epidemiology. 2018;47(6):2005-14.

\bibitem{hill2011bayesian}
Hill JL.
\newblock Bayesian nonparametric modeling for causal inference.
\newblock Journal of Computational and Graphical Statistics. 2011;20(1):217-40.

\bibitem{wager2018estimation}
Wager S, Athey S.
\newblock Estimation and inference of heterogeneous treatment effects using
  random forests.
\newblock Journal of the American Statistical Association.
  2018;113(523):1228-42.

\bibitem{johansson2016learning}
Johansson F, Shalit U, Sontag D.
\newblock Learning representations for counterfactual inference.
\newblock In: International conference on machine learning; 2016. p. 3020-9.

\bibitem{pmlr-v70-shalit17a}
Shalit U, Johansson FD, Sontag D.
\newblock Estimating individual treatment effect: generalization bounds and
  algorithms.
\newblock In: Precup D, Teh YW, editors. Proceedings of the 34th International
  Conference on Machine Learning. vol.~70 of Proceedings of Machine Learning
  Research. International Convention Centre, Sydney, Australia: PMLR; 2017. p.
  3076-85.
\newblock Available from:
  \url{http://proceedings.mlr.press/v70/shalit17a.html}.

\bibitem{shi2019adapting}
Shi C, Blei D, Veitch V.
\newblock Adapting neural networks for the estimation of treatment effects.
\newblock In: Advances in neural information processing systems; 2019. p.
  2507-17.

\bibitem{alaa2017deep}
Alaa AM, Weisz M, Van Der~Schaar M.
\newblock Deep counterfactual networks with propensity-dropout.
\newblock arXiv preprint arXiv:170605966. 2017.

\bibitem{Yoon2018}
Yoon J, Jordon J, Van Der~Schaar M.
\newblock GANITE: Estimation of individualized treatment effects using
  generative adversarial nets.
\newblock In: International Conference on Learning Representations; 2018. .

\bibitem{louizos2017causal}
Louizos C, Shalit U, Mooij JM, Sontag D, Zemel R, Welling M.
\newblock Causal effect inference with deep latent-variable models.
\newblock In: Advances in neural information processing systems; 2017. p.
  6446-56.

\bibitem{zhang2020treatment}
Zhang W, Liu L, Li J.
\newblock Treatment effect estimation with disentangled latent factors.
\newblock arXiv preprint arXiv:200110652. 2020.

\bibitem{makhzani2015adversarial}
Makhzani A, Shlens J, Jaitly N, Goodfellow I, Frey B.
\newblock Adversarial autoencoders.
\newblock arXiv preprint arXiv:151105644. 2015.

\bibitem{kingma2013auto}
Kingma DP, Welling M.
\newblock Auto-encoding variational bayes.
\newblock arXiv preprint arXiv:13126114. 2013.

\bibitem{goodfellow2014generative}
Goodfellow I, Pouget-Abadie J, Mirza M, Xu B, Warde-Farley D, Ozair S, et~al.
\newblock Generative adversarial nets.
\newblock Advances in neural information processing systems. 2014;27:2672-80.

\bibitem{chen2016infogan}
Chen X, Duan Y, Houthooft R, Schulman J, Sutskever I, Abbeel P.
\newblock Infogan: Interpretable representation learning by information
  maximizing generative adversarial nets.
\newblock Advances in neural information processing systems. 2016;29:2172-80.

\bibitem{funk2011doubly}
Funk MJ, Westreich D, Wiesen C, St{\"u}rmer T, Brookhart MA, Davidian M.
\newblock Doubly robust estimation of causal effects.
\newblock American journal of epidemiology. 2011;173(7):761-7.

\bibitem{dudik2014doubly}
Dud{\'\i}k M, Erhan D, Langford J, Li L, et~al.
\newblock Doubly robust policy evaluation and optimization.
\newblock Statistical Science. 2014;29(4):485-511.

\bibitem{rubin1974}
Rubin DB.
\newblock Estimating causal effects of treatments in randomized and
  nonrandomized studies.
\newblock Journal of educational Psychology. 1974;66(5):688-701.

\bibitem{rosenbaum1983}
Rosenbaum PR, Rubin DB.
\newblock {The central role of the propensity score in observational studies
  for causal effects}.
\newblock Biometrika. 1983 04;70(1):41-55.

\bibitem{imbens2000role}
Imbens GW.
\newblock The role of the propensity score in estimating dose-response
  functions.
\newblock Biometrika. 2000;87(3):706-10.

\bibitem{pearl2016causal}
Pearl J, Glymour M, Jewell NP.
\newblock Causal Inference in Statistics: A Primer.
\newblock Wiley; 2016.
\newblock Available from: \url{https://books.google.com/books?id=L3G-CgAAQBAJ}.

\bibitem{porter2011relative}
Porter KE, Gruber S, Van Der~Laan MJ, Sekhon JS.
\newblock The relative performance of targeted maximum likelihood estimators.
\newblock The International Journal of Biostatistics. 2011;7(1).

\bibitem{chipman2010bart}
Chipman HA, George EI, McCulloch RE, et~al.
\newblock BART: Bayesian additive regression trees.
\newblock The Annals of Applied Statistics. 2010;4(1):266-98.

\bibitem{athey2016recursive}
Athey S, Imbens G.
\newblock Recursive partitioning for heterogeneous causal effects.
\newblock Proceedings of the National Academy of Sciences.
  2016;113(27):7353-60.

\bibitem{lu2018estimating}
Lu M, Sadiq S, Feaster DJ, Ishwaran H.
\newblock Estimating individual treatment effect in observational data using
  random forest methods.
\newblock Journal of Computational and Graphical Statistics. 2018;27(1):209-19.

\bibitem{dehejia2002propensity}
Dehejia RH, Wahba S.
\newblock Propensity score-matching methods for nonexperimental causal studies.
\newblock Review of Economics and statistics. 2002;84(1):151-61.

\bibitem{lunceford2004stratification}
Lunceford JK, Davidian M.
\newblock Stratification and weighting via the propensity score in estimation
  of causal treatment effects: a comparative study.
\newblock Statistics in medicine. 2004;23(19):2937-60.

\bibitem{crump2008nonparametric}
Crump RK, Hotz VJ, Imbens GW, Mitnik OA.
\newblock Nonparametric tests for treatment effect heterogeneity.
\newblock The Review of Economics and Statistics. 2008;90(3):389-405.

\bibitem{pmlr-v80-alaa18a}
Alaa A, van~der Schaar M.
\newblock Limits of Estimating Heterogeneous Treatment Effects: Guidelines for
  Practical Algorithm Design.
\newblock In: Dy J, Krause A, editors. Proceedings of the 35th International
  Conference on Machine Learning. vol.~80 of Proceedings of Machine Learning
  Research. Stockholmsmässan, Stockholm Sweden: PMLR; 2018. p. 129-38.
\newblock Available from: \url{http://proceedings.mlr.press/v80/alaa18a.html}.

\bibitem{blei2017variational}
Blei DM, Kucukelbir A, McAuliffe JD.
\newblock Variational inference: A review for statisticians.
\newblock Journal of the American statistical Association.
  2017;112(518):859-77.

\bibitem{almond2005costs}
Almond D, Chay KY, Lee DS.
\newblock The costs of low birth weight.
\newblock The Quarterly Journal of Economics. 2005;120(3):1031-83.

\bibitem{lalonde1986evaluating}
LaLonde RJ.
\newblock Evaluating the econometric evaluations of training programs with
  experimental data.
\newblock The American economic review. 1986:604-20.

\bibitem{kingma2014adam}
Kingma DP, Ba J.
\newblock Adam: A method for stochastic optimization.
\newblock arXiv preprint arXiv:14126980. 2014.

\end{thebibliography}

\section*{Appendix}
\subsection*{A1 Algorithms}
The algorithm to train the generative adversarial network and doubly robust multitask network for counterfactual outcome calculation and ITE estimation are discussed in Algorithms \ref{algo: CF_GM} and \ref{algo: DR_model} respectively.
\subsection*{A2 Performance Metrics}
\label{pmetrics}
The error for PEHE, ATE, Policy Risk, ATT will be evaluated by estimating $\epsilon_{PEHE}, \epsilon_{ATE}, R_{pol}(\pi), \epsilon_{ATT}$ respectively as follows:
\begin{align} 
    &\epsilon_{PEHE} = \frac{1}{N}\sum_{n=0}^N\Big(\mathbb{E}_{y_{j}(n)\sim \mu_j(n)}\big[y_1(n) - y_0(n)\big] 
    - \big[\hat{y_1}(n) - \hat{y_0}(n)\big]\Big)^2 \\
    &\epsilon_{ATE} = ||\frac{1}{N}\sum_{n=0}^N\mathbb{E}_{y(n) \sim \mu(n)}[y(n)] 
    -\frac{1}{N} \sum_{n=0}^N\hat y(n) ||_2^2 \\
    &R_{pol}(\pi) = \frac{1}{N}\sum_{n=0}^N\Big[1 - 
    \Big(\sum_{i=1}^{k}\Big[\frac{1}{|{\Pi_i\cap T_i \cap E}|} 
    \sum_{x(n)\in \Pi_i\cap\ T_i\cap E} y_i(n) \times \frac{|\Pi_n \cap E|}{|E|}\Big] \Big)\Big] 
\label{equ: PEHE}
\end{align}

where $\pi_i$ = $\{\boldsymbol{x(n)}: i$ = arg max $\hat{\boldsymbol{y}}$ , 
\\$T_i = {\boldsymbol{x(n)} : t_i(n) = 1\}}$, and $E$ is the randomized sample.

The true average treatment effect on the treated (ATT) and its error $\epsilon_{ATT}$ are defined as follows:

\begin{align} 
\label{ATT}
    &ATT = \frac{1}{|T_1 \cap E|} \sum_{x_i \in T_1 \cap E} Y_1(x_i) - 
    \frac{1}{|T_0 \cap E|} \sum_{x_i \in C \cap E} Y_0(x_i) \\
    &\epsilon_{ATT} =\big|ATT - \frac{1}{|T_1 \cap E|}\sum_{x_i \in T_1 \cap E} 
    \hat{Y}_1(x_i) - \hat{Y}_0(x_i)\big|
\end{align}
\label{equ: ATT}
where $T_1$, $T_0$ and $E$ are the subsets corresponding to treated, controlled samples, and randomized controlled trials, respectively. 

\subsection*{A3 Synthetic dataset of CEVAE} 
\label{CEVAE_dataset}
\begin{align}
\textbf{z}_i \sim \text{Bern}(0.5); & \qquad
\textbf{x}_i|\textbf{z}_i \sim \mathcal{N}(\textbf{z}_i, \sigma^2_{5}\textbf{z}_i + \sigma^2_{3}(1 - \textbf{z}_i)) \nonumber\\
t_i|\textbf{z}_i &\sim \text{Bern}(0.75\textbf{z}_i + 0.25(1 - \textbf{z}_i)) \nonumber\\
\textbf{y}_i|t_i, \textbf{z}_i &\sim \text{Bern}(\text{Sigmoid}(3(\textbf{z}_i + 2(2t_i - 1))))
\label{equ: CEVAE_data}
\end{align}

\subsection*{A4 Synthetic dataset of DR-VIDAL} 
\label{GANITE_dataset}
\begin{align}
\textbf{z}_x \sim \text{Bern}(0.5); & \qquad \textbf{z}_t \sim \text{Bern}(0.5) \nonumber\\
\textbf{z}_{yf} \sim \text{Bern}(0.5); & \qquad \textbf{z}_{ycf} \sim \text{Bern}(0.5) \nonumber\\
\textbf{x}_x|\textbf{z}_x &\sim \mathcal{N}(\textbf{z}_x, 5(\textbf{z}_x) + 3(1 - \textbf{z}_x)) \nonumber\\
\textbf{x}_t|\textbf{z}_t &\sim \mathcal{N}(\textbf{z}_x, 2(\textbf{z}z_t) + 0.5(1 - \textbf{z}_t)) \nonumber\\
\textbf{x}_{yf}|\textbf{z}_{yf} &\sim \mathcal{N}(\textbf{z}_{yf}, 10(\textbf{z}_{yf}) + 6(1 - \textbf{z}_{yf})) \nonumber\\
\textbf{x}_{ycf}|\textbf{z}_{ycf} &\sim \mathcal{N}(\textbf{z}_{ycf}, 10(\textbf{z}_{ycf}) + 6(1 - \textbf{z}_{ycf})) \nonumber\\
\textbf{w}_t^T \sim \mathcal{U}((-0.1, 0.1)^{10 \text{x} 1}); & \qquad
\textbf{n}_t \sim \mathcal{N}(0, 0.1)\nonumber\\
\textbf{w}_y^T \sim \mathcal{U}((-1, 1)^{10 \text{x} 2}); & \qquad 
\textbf{n}_y \sim \mathcal{N}(0^{2 \text{x} 1}, 0.1 \text{x} \mathcal{I}^{2 \text{x} 2})\nonumber\\
t|x \sim \text{Bern}(\text{Sigmoid}(\textbf{w}_t^T\textbf{x} + \textbf{n}_t)); & \qquad
\textbf{y}|\textbf{x} \sim \textbf{w}_y^T\textbf{x} + \textbf{n}_y
\label{equ: DR_VIDAL_data}
\end{align}

\subsection*{A5 Datasets}
\label{A-Datasets}
The IHDP and Twins two are semi-synthetic, and simulated counterfactuals to the real factual data are available. These datasets have been also designed and collated to meet specific treatment overlap condition, nonparallel treatment assignment, and nonlinear outcome surfaces \cite{hill2011bayesian, pmlr-v70-shalit17a, louizos2017causal, Yoon2018}. The IHDP datasets is composed by 110 treated subjects and 487 controls, with 25 covariates. The Twins dataset comprises 4553 treated, 4567 controls, with 30 covariates. The Jobs dataset comprises 237 treated, 2333 controls, with 17 covariates. For all the real-world datasets, we use the same experimental settings described in GANITE, where the datasets are divided into 56/24/20 \% train-validation-test splits. We run 1000, 10 and 100 realizations of IHDP, Jobs and Twins datasets, respectively. 

\subsection*{A7 Differences with CEVAE and GANITE}
The counterfactual outcome predictor of DR-VIDAL uses both VAE and GAN in the same framework, while only VAE is used in CEVAE and only GAN is used GANITE. CEVAE also incorporates a causal graph, but it is simplistic, as it infers only the observed proxy $X$ from $Z$. We instead considered multiple latent variables causally related to the treatment and the outcome in addition to the direct links to the pre-treatment covariates. Furthermore, we use GAN to generate counterfactual examples, but, unlike GANITE, we first infer the multiple latent factors using a VAE, then optimize the GAN with the mutual information, and finally generate the entire potential outcome vector.

\subsection*{A8 Differences with TARNet and Dragonnet}
The design of the doubly robust module block of DR-VIDAL is closely related to that of TARNet and Dragonnet. However, TARNet uses a two-headed network, which is not doubly robust. Dragonnet includes a third head that incorporates the propensity score. DR-VIDAL exploits the doubly robustness adding two heads, i.e., the propensity score and the regressor head, to the the basic two-headed TARNet configuration. Further, in TARNet the weights corresponding to each sample are calculated as the crude probability of the treatment assignment, whereas DR-VIDAL accounts for the pre-treatment covariates. For Dragonnet, the targeted regularization is implemented without taking into account the regressed outcome, which instead is estimated by DR-VIDAL in the fourth head, as a function of treatment and pre-treatment covariates. Another major difference between TARNet/Dragonnet and DR-VIDAL is the training strategy. For both TARNet and Dragonnet, the counterfactual outcome does not exist, so for each sample the overall loss function has to be estimated with the factual outcome only, updating the parameters of the outcome head of the factual outcome during training. In contrast DR-VIDAL provides the entire potential outcome vector, comprising both the factual and the counterfactual outcomes. For each training sample, the loss function is calculated for both outcomes, and the corresponding parameters of both the outcome heads are updated.

\subsection*{A6 Derivation of the ELBO Loss for VAE} 
\label{ELBO_Loss_derivation}

From Figure section 4.1 and the causal graph in figure 1 in the main text, $p_{\phi_d}(\textbf{x}|\textbf{z}_x, \textbf{z}_t, \textbf{z}_{yf}, \textbf{z}_{y_{cf}})$ and 
$p_{\phi_d}(\textbf{z}_x, \textbf{z}_t, \textbf{z}_{yf}, \textbf{z}_{y_cf}|\textbf{x})$ are the true likelihood and true posterior respectively. The posterior is hard to evaluate, so we have to approximate the true posterior to the product of the factorized known distributions $q_{\phi_x}(\textbf{z}_x|\textbf{x})$, $q_{\phi_t}(\textbf{z}_t|\textbf{x})$, $q_{\phi_{yf}}(\textbf{z}_{yf}|\textbf{x})$ and $q_{\phi_{ycf}}(\textbf{z}_{ycf}|\textbf{x})$ by minimising the KL divergence as follows,

\begin{align*}
&KL\big(q_{\phi_x}(\textbf{z}_x|\textbf{x})q_{\phi_t}(\textbf{z}_t|\textbf{x})q_{\phi_{yf}}(\textbf{z}_{yf}|\textbf{x})q_{\phi_{ycf}}(\textbf{z}_{ycf}|\textbf{x})||
p_{\phi_d}(\textbf{z}_x, \textbf{z}_t, \textbf{z}_{yf}, \textbf{z}_{ycf}|\textbf{x})\big) \\
    &= \int\int\int\int q_{\phi_x}(\textbf{z}_x|\textbf{x})q_{\phi_t}(\textbf{z}_t|\textbf{x})q_{\phi_{yf}}(\textbf{z}_{yf}|\textbf{x})q_{\phi_{ycf}}(\textbf{z}_{ycf}|\textbf{x})
    \big[\log \frac{q_{\phi_x}(\textbf{z}_x|\textbf{x})q_{\phi_t}(\textbf{z}_t|\textbf{x})q_{\phi_{yf}}(\textbf{z}_{yf}|x)}{p_{\phi_d}(\textbf{z}_x, \textbf{z}_t, \textbf{z}_{yf}, \textbf{z}_{ycf}|\textbf{x})}  \big] \textbf{dz}_x \textbf{dz}_t \textbf{dz}_{yf} \textbf{dz}_{ycf} \\
% \end{align*}
% \begin{align*}
    &= \int\int\int\int q_{\phi_x}(\textbf{z}_x|\textbf{x})q_{\phi_t}(\textbf{z}_t|\textbf{x})q_{\phi_{yf}}(\textbf{z}_{yf}|\textbf{x})q_{\phi_{ycf}}(\textbf{z}_{ycf}|\textbf{x}) \\
    &\qquad\big[\log q_{\phi_x}(\textbf{z}_x|\textbf{x}) + \log q_{\phi_t}(\textbf{z}_t|\textbf{x})  
    + \log q_{\phi_{yf}}(\textbf{z}_{yf}|\textbf{x}) + \log q_{\phi_{ycf}}(\textbf{z}_{ycf}|\textbf{x}) 
    - \log p_{\phi_d}(\textbf{z}_x, \textbf{z}_t, \textbf{z}_{yf}, \textbf{z}_{ycf}|\textbf{x})  \big]
    \textbf{dz}_x \textbf{dz}_t \textbf{dz}_{yf} \textbf{dz}_{ycf}\\
% \end{align*}
% \begin{align*}
    &= \int\int\int\int q_{\phi_x}(\textbf{z}_x|\textbf{x})q_{\phi_t}(\textbf{z}_t|\textbf{x})q_{\phi_{yf}}(\textbf{z}_{yf}|\textbf{x})q_{\phi_{ycf}}(\textbf{z}_{ycf}|\textbf{x})
    \big[\log q_{\phi_x}(\textbf{z}_x|\textbf{x}) + \log q_{\phi_t}(\textbf{z}_t|\textbf{x}) + \\
    & \qquad \log q_{\phi_{yf}}(\textbf{z}_{yf}|\textbf{x}) + \log q_{\phi_{ycf}}(\textbf{z}_{ycf}|\textbf{x})
    - \log p_{\phi_d}(x|\textbf{z}_x, \textbf{z}_t, \textbf{z}_{yf}, \textbf{z}_{ycf}) - \log p_{\phi_d}(\textbf{z}_x, \textbf{z}_t, \textbf{z}_{yf}, \textbf{z}_{ycf}) \\
    &\qquad+ \log{p_{\phi_d}(x)}   \big]
    \textbf{dz}_x \textbf{dz}_t \textbf{dz}_{yf} \textbf{dz}_{ycf}\\
% \end{align*}
% \begin{align*}
    &= \int q_{\phi_x}(\textbf{z}_x|\textbf{x}) \log \frac{q_{\phi_x}(\textbf{z}_x|\textbf{x})}{p_{\phi_d}(\textbf{z}_x)}\textbf{dz}_x 
    + \int q_{\phi_t}(\textbf{z}_t|\textbf{x}) \log \frac{q_{\phi_t}(\textbf{z}_t|\textbf{x})}{p_{\phi_d}(\textbf{z}_t)}\textbf{dz}_t
    + \int q_{\phi_{yf}}(\textbf{z}_{yf}|\textbf{x}) \log \frac{q_{\phi_{yf}}(\textbf{z}_{yf}|\textbf{x})}{p_{\phi_d}(\textbf{z}_{yf})}\textbf{dz}_{yf} \\
    &\qquad+ \int q_{\phi_{ycf}}(\textbf{z}_{ycf}|\textbf{x}) \log \frac{q_{\phi_x}(\textbf{z}_{ycf}|\textbf{x})}{p_{\phi_d}(\textbf{z}_x)}\textbf{dz}_{ycf}
    -\int\int\int\int \big[
    q_{\phi_x}(\textbf{z}_x|\textbf{x})q_{\phi_t}(\textbf{z}_t|\textbf{x})q_{\phi_{yf}}(\textbf{z}_{yf}|\textbf{x}) \\
    &\qquad q_{\phi_{ycf}}(\textbf{z}_{ycf}|\textbf{x})\log p_{\phi_d}(\textbf{x}|\textbf{z}_x, \textbf{z}_t, \textbf{z}_{yf}, \textbf{z}_{ycf})\big]
    \textbf{dz}_x \textbf{dz}_t \textbf{dz}_{yf} \textbf{dz}_{ycf}
    + \log p_{\phi_d}(\textbf{x}) \\
% \end{align*}
% \begin{align*}
    &=KL\big(q_{\phi_x}(\textbf{z}_x|\textbf{x}) || p_{\phi_d}(\textbf{z}_x))\big) 
     + KL\big(q_{\phi_t}(\textbf{z}_t|\textbf{x}) || p_{\phi_d}(\textbf{z}_t))\big) +KL\big(q_{\phi_{yf}}(\textbf{z}_{yf}|\textbf{x}) || p_{\phi_d}(\textbf{z}_{yf}))\big) \\
    & \qquad + KL\big(q_{\phi_{ycf}}(\textbf{z}_{ycf}|\textbf{x}) || p_{\phi_d}(\textbf{z}_{ycf}))\big) -\mathbb{E}_{q_{\phi_x}, q_{\phi_t}, q_{\phi_{yf}}, q_{\phi_{ycf}}}[\log p(\textbf{x}|\textbf{z}_x, \textbf{z}_t, \textbf{z}_{yf}, \textbf{z}_{ycf})]\\
    & \qquad + \log p_{\phi_d}(\textbf{x})
\end{align*}

where, the distributions $q_{\phi_x}(\textbf{z}_x|x), q_{\phi_t}(\textbf{z}_t|x)$, $q_{\phi_{yf}}(z_{yf}|x)$, 
$q_{\phi_{ycf}}(z_{ycf}|x)$ and $p_{\phi_d}(x|\textbf{z}_x, \textbf{z}_t, \textbf{z}_{yf}, \textbf{z}_{ycf})$ are parameterized by the parameters
$\phi_x, \phi_t, \phi_{yf}, \phi_{ycf}, \phi_d$. The KL divergence of two distributions is always greater than or equal to zero. So,

\begin{align}
 & KL\big(q_{\phi_x}(\textbf{z}_x|x)q_{\phi_t}(\textbf{z}_t|x)q_{\phi_{yf}}(\textbf{z}_{yf}|\textbf{x})q_{\phi_{ycf}}(\textbf{z}_{ycf}|x)||
 p_{\phi_d}(\textbf{z}_x, \textbf{z}_t, \textbf{z}_{yf}, \textbf{z}_{ycf}|x)\big) \geq 0, \nonumber\\
&\log p_{\phi_d}(\textbf{z}) \geq \mathcal{L}_{ELBO} \quad \text{where,} \nonumber\\ 
 &\mathcal{L}_{ELBO}(\phi_x, \phi_t, \phi_{yf}, \phi_{ycf}; \textbf{x}, \textbf{z}_x, \textbf{z}_t, \textbf{z}_{yf}, \textbf{z}_{ycf}) \nonumber\\
&\quad= \mathbb{E}_{q_{\phi_x}, q_{\phi_t}, q_{\phi_{yf}}, q_{\phi_{ycf}}}[\log p(\textbf{\textbf{x}}|\textbf{z}_x, \textbf{z}_t, \textbf{z}_{yf}, \textbf{z}_{ycf})]
\nonumber\\
 &\qquad -KL\big(q_{\phi_x}(\textbf{z}_x|\textbf{x}) || p_{\phi_d}(\textbf{z}_x))\big) - KL\big(q_{\phi_t}(\textbf{z}_t|\textbf{x}) || p_{\phi_d}(\textbf{z}_t))\big) \nonumber - KL\big(q_{\phi_{yf}}(\textbf{z}_{yf}|\textbf{x}) || p_{\phi_d}(\textbf{z}_{yf}))\big) \nonumber\\
     &\qquad - KL\big(q_{\phi_{ycf}}(\textbf{z}_{ycf}|\textbf{x}) || p_{\phi_d}(\textbf{z}_{ycf}))\big) \nonumber
\end{align}

\begin{algorithm}
	\caption{Training of the generative adversarial network for counterfactual outcome calculation} 
	\label{algo: CF_GM}
	\hspace*{\algorithmicindent} \textbf{Input:}  {Training set $\boldsymbol{X}$ = \{($\textbf{x}^{(1)}$, $t^{(1)}$, $y_f^{(1)}$),...,
	($\textbf{x}^{(n)}$, $t^{(n)}$, $y_f^{(n)}$) \}; hyper-parameters $\gamma > $  0; $\lambda > $  0; Encoders: $E_{\phi_x}$,
	$E_{\phi_t}$, $E_{\phi_{yf}}$, $E_{\phi_{ycf}}$ with parameters ${\phi_x}, {\phi_t}, {\phi_{yf}}, {\phi_{ycf}}$ respectively; Decoder $D_{\phi_d}$ with parameter $D_{\phi_d}$; Generator $G_{\theta_g}$, Discriminator $D_{\theta_d}$, Q network 
	$D_{\theta_q}$ with parameters ${\theta_g}, {\theta_d}, {\theta_q}$ respectively}
	\begin{algorithmic}[1]
        \State Initialize parameters: ${\phi_x}, {\phi_t}, {\phi_{yf}}, {\phi_{ycf}}, {\phi_d}, {\theta_g}, {\theta_d}, {\theta_q}$
        \While {training}
            \State $\textbf{x} \leftarrow$ batch of samples from the dataset  
            \State $\textbf{z}_{\mu_x}, \textbf{z}_{\sigma_x}\leftarrow$ $E_{\phi_x}(\textbf{x})$ 
            \State $\textbf{z}_{\mu_t}, \textbf{z}_{\sigma_t}\leftarrow$ $E_{\phi_t}(\textbf{x})$
            \State $\textbf{z}_{\mu_{yf}}, \textbf{z}_{\sigma_{yf}}\leftarrow$ $E_{\phi_{yf}}(\textbf{x})$
            \State $\textbf{z}_{\mu_{ycf}}, \textbf{z}_{\sigma_{ycf}}\leftarrow$ $E_{\phi_{ycf}}(\textbf{x})$
            \State $\textbf{z}_{x} \leftarrow \textbf{z}_{\mu_x} + \epsilon \textbf{z}_{\sigma_x}$ , where $\epsilon \sim \mathcal{N}(0, Id)$
            \State $\textbf{z}_{t} \leftarrow \textbf{z}_{\mu_t} + \epsilon \textbf{z}_{\sigma_t}$ , where $\epsilon \sim \mathcal{N}(0, Id)$
            \State $\textbf{z}_{yf} \leftarrow \textbf{z}_{\mu_{yf}} + \epsilon \textbf{z}_{\sigma_{yf}}$ , where $\epsilon \sim \mathcal{N}(0, Id)$
            \State $\textbf{z}_{ycf} \leftarrow \textbf{z}_{\mu_{ycf}} + \epsilon \textbf{z}_{\sigma_{ycf}}$ , where $\epsilon \sim \mathcal{N}(0, Id)$
            \State Concatenate $\textbf{z}_{x}, \textbf{z}_{t}, \textbf{z}_{yf}, \textbf{z}_{ycf}$ to form $\textbf{z}_c$
            \State $\hat{\textbf{x}}\leftarrow D_{\phi_d}(\textbf{z}_c)$
            \State Calculate $\mathcal{L}_{VAE}(\phi_x, \phi_t, \phi_{yf}, \phi_{ycf}; \textbf{x}, \textbf{z}_x, \textbf{z}_t, \textbf{z}_{yf}, \textbf{z}_{ycf})$
            \State $\phi_x \xleftarrow{-} \nabla_{\phi_x}\mathcal{L}_{VAE}$; 
            $\phi_t \xleftarrow{-} \nabla_{\phi_t}\mathcal{L}_{VAE}$;
            $\phi_{yf} \xleftarrow{-} \nabla_{\phi_{yf}}\mathcal{L}_{VAE}$;
            \hspace*{\algorithmicindent} $\phi_{ycf} \xleftarrow{-} \nabla_{\phi_{ycf}}\mathcal{L}_{VAE}$;
            $\phi_d \xleftarrow{-} \nabla_{\phi_d}\mathcal{L}_{VAE}$
            \State  $\textbf{z}_G \sim \mathcal{N}(0,Id)$
            \State $y_0, y_1 \leftarrow G_{\theta_g}(\textbf{z}_G, \textbf{z}_c)$
            \State $\hat{y}_0 = ((1 - t) * y_f + t * y_0)$
            ; $\hat{y}_1 = (t * y_f + (1 - t) * y_1)$
            \State $d_{logit} \leftarrow D_{\theta_d}(\textbf{x}, \hat{y}_0, \hat{y}_1)$
            \State Calculate $\mathcal{L}^D(\theta_d)$
            \State $\theta_d \xleftarrow{-} \nabla_{\theta_d} \mathcal{L}^D(\theta_d)$
            \State $\hat{y}_f \leftarrow t * y_1 + (1 - t) * y_0$
            \State Compute$\mathcal{L}^G_S(y_f, \hat{y}_f)$
            \State Concatenate $y_0, y_1$ to form $q_{input}$
            \State $q_\mu, q_\sigma \leftarrow Q_{\theta_q}(q_{input})$
            \State Compute $\mathcal{L}_I(G, Q)$ by treating $Q(c|x)$ as factored Gaussian 
            using $q_\mu, q_\sigma$ and $z_c$
            \State Compute $\mathcal{L}^G(\theta_g)$
            \State $\theta_g \xleftarrow{-} \nabla_{\theta_g} \mathcal{L}^G(\theta_g)$
        \EndWhile
	\end{algorithmic} 
\end{algorithm}

\begin{algorithm}[t]
	\caption{Training of the doubly robust multitask network for ITE estimation} 
	\label{algo: DR_model}
	\hspace*{\algorithmicindent} \textbf{Input:}  {Complete dataset \textbf{$\Tilde{X}$} = \{($\textbf{x}^{(1)}$, $t^{(1)}$, $y_f^{(1)}$, $y_{cf}^{(1)}$),...,
	($\textbf{x}^{(n)}$, $t^{(n)}$, $y_f^{(n)}$, $y_{cf}^{(n)}$) \} after training the GAN module for counterfactual prediction; hyper-parameters $\alpha > $  0; $\beta > $  0; outcome heads with shared parameters $\phi$ and outcome specific parameters $\theta_0, \theta_1$; propensity head with parameters $\theta_\pi$; regressor head with parameters $\theta_\mu$}
	\begin{algorithmic}[1]
        \State Initialize parameters: $\theta_0, \theta_1, \theta_\pi, \theta_\mu$
        \While {training}
            \State $\textbf{x} \leftarrow$ batch of samples from the dataset  
            \State Calculate $\hat{y}_i^{(0)}, \hat{y}_i^{(1)}, \hat{y}_f^{(i)}, \hat{y}_{cf}^{(i)}$
            \State Calculate the predicted loss $\mathcal{L}_i^p(\theta_1, \theta_0, \phi)$
            \State Calculate $\hat{y}_{fDR}^{(i)}, \hat{y}_{cfDR}^{(i)}$
            \State Calculate the doubly Robust loss $\mathcal{L}_i^{DR}(\theta_1, \theta_0, \theta_\pi, \theta_\mu, \phi)$
            \State Calculate the final loss $\mathcal{L}_{ITE}(\theta_1, \theta_0, \theta_\pi, \theta_\mu, \phi)$
            \State Calculate gradients of the loss $\mathcal{L}_{ITE}(\theta_1, \theta_0, \theta_\pi, \theta_\mu, \phi)$
            \State Update the parameters $\theta_1, \theta_0, \theta_\pi, \theta_\mu, \phi$
        \EndWhile
        
	\end{algorithmic} 
\end{algorithm}

\begin{table}[h]
\fontsize{7.2pt}{0.35cm}\selectfont
\centering
\begin{tabular}{l|l|l|l}

\toprule 
      & \multicolumn{1}{c}{\textbf{IHDP}} & \multicolumn{1}{|c}{\textbf{Jobs}} & \multicolumn{1}{|c}{\textbf{Twins}} \\
      & $\boldsymbol{\sqrt{\epsilon^{out-of-s}_{PEHE}}}$ 
      & $\boldsymbol{R^{out-of-s}_{Pol}}$ 
      & $\boldsymbol{\sqrt{\epsilon^{out-of-s}_{PEHE}}}$  \\
\midrule 
      \textbf{DR-VIDAL}      & \textbf{0.62 $\pm$ 0.06}  & \textbf{0.102 $\pm$ 0.01} & 
       \textbf{0.318 $\pm$ 0.008} \\
      DR-VIDAL (w/o DR loss) & 0.85 $\pm$ 0.06 & 0.110 $\pm$ 0.01  &
      0.324 $\pm$ 0.007 \\
      DR-VIDAL (w/o Info loss)    & 0.67 $\pm$ 0.04 & 0.109 $\pm$ 0.01 &
      0.318 $\pm$ 0.012\\
      DR-VIDAL (w/o DR + Info loss) & 0.81 $\pm$ 0.05 & 0.113 $\pm$ 0.01 &
      0.326 $\pm$ 0.008 \\
\bottomrule
\end{tabular}
\caption{Performance of the all the different DR-VIDAL configurations on the IHDP, Jobs and Twins datasets (1000, 10, and 100 realizations, respectively). Results show the out-of-sample (mean $\pm$ st.dev) error (PEHE) and policy risk ($R_{Pol}$).}\smallskip
\label{tab: big_table}
\end{table}

\subsection*{A9 Training and implementation of DR-VIDAL}
\label{A-Training}
\paragraph{Adversarial module.} To reduce the model complexity and parameters for the encoder of the VAE, we have a shared neural network connected to 4 other networks for estimating the four posterior distributions $q_{\phi_x}(\textbf{z}_x|\textbf{x})$, $q_{\phi_t}(\textbf{z}_t|\textbf{x})$, $q_{\phi_{yf}}(\textbf{z}_{yf}|\textbf{x})$, $q_{\phi_{ycf}}(\textbf{z}_{ycf}|\textbf{x})$. The shared neural network has 3 layers, each with 15 nodes. The networks with $q_{\phi_x}(\textbf{z}_x|\textbf{x})$, $q_{\phi_t}(\textbf{z}_t|\textbf{x})$, $q_{\phi_{yf}}(\textbf{z}_{yf}|x)$, $q_{\phi_{ycf}}(\textbf{z}_{ycf}|\textbf{x})$ as outputs have a single layer with 5, 1, 1, 1 nodes, respectively. The decoder is a 4-layer neural network, each with 15 nodes to calculate the data likelihood $p_{\phi_d}(\textbf{x}|\textbf{z}_x, \textbf{z}_t, \textbf{z}_{yf}, \textbf{z}_{ycf})$. For the GAN, the generator network has 2 shared layers and 2 outcome-specific layers, each with 100 nodes. The discriminator and the network for information maximization (Q network in Figure \ref{fig: vae_gan}) is a 3-layered neural network, each with 30 nodes and 8 nodes respectively. All the layers of the VAE and GAN use Rectified Linear Unit (ReLU) activation functions and the parameters are updated using the Adam optimizer \cite{kingma2014adam}. The random noise $\textbf{z}_G$ is sampled from a 92-dimensional standardized Gaussian distribution $\mathcal{N}(0,1)$. The hyperparameter $\gamma$ is set as 1 for all datasets, while $\lambda$ is set as 0.2, 0.01 and 10 for IHDP, Jobs and Twins, respectively. The batch sizes of IHDP, Jobs, and Twins are 64, 64, and 256, respectively. The learning rates of the VAE, generator and discriminator are 1e-3, 1e-4, and 5e-4, respectively.

\begin{table} [h]
\fontsize{7.6pt}{0.35cm}\selectfont
% \small
\centering
\begin{tabular}{l|l r}
\toprule 
    %   & \multicolumn{2}{c}{\textbf{IHDP}} & \multicolumn{2}{|c}{\textbf{Jobs}} \\
      $\boldsymbol{Methods}$ & Out-Sample & In-Sample \\
\midrule 
      OLS/LR1   & 0.08 $\pm$ 0.04  & 0.01 $\pm$ 0.00  \\
      OLS/LR2   & 0.08 $\pm$ 0.03 & 0.01 $\pm$ 0.01   \\
      BLR       & 0.08 $\pm$ 0.03 & 0.01 $\pm$ 0.011  \\
      k-NN      & 0.13 $\pm$ 0.05 & 0.21 $\pm$ 0.01   \\
\midrule 
      BART      & 0.08 $\pm$ 0.03 & 0.02 $\pm$ 0.00 \\
      R Forest  & 0.09 $\pm$ 0.04 & 0.03 $\pm$ 0.01 \\
      C Forest  & 0.07 $\pm$ 0.03 & 0.03 $\pm$ 0.01 \\
\midrule 
      BNN      & 0.09 $\pm$ 0.04 & 0.03 $\pm$ 0.01 \\
      TARNET    & 0.11 $\pm$ 0.04 & 0.05 $\pm$ 0.02 \\
      CFR$_{WASS}$ & 0.09 $\pm$ 0.03 & 0.04 $\pm$ 0.01 \\
\midrule 
      GANITE   & 0.06 $\pm$ 0.03 & 0.01 $\pm$ 0.01 \\
      CEVAE    &0.03 $\pm$ 0.01 & 0.02 $\pm$ 0.01 \\
\midrule
      \textbf{DR-VIDAL}  & \textbf{0.05 $\pm$ 0.02} & \textbf{0.04 $\pm$ 0.03}  \\
\bottomrule
\end{tabular}
\caption{Performance of 
various models on the Jobs dataset for $\epsilon_{ATT}$ (mean $\pm$ st.dev).}\smallskip
\label{tab: Jobs_within}
\end{table}

\paragraph{Doubly robust module.} For the doubly robust module, the shared network $f_\phi$ and outcome specific networks $f_{\theta_0}$ and $f_{\theta_1}$ are both 3-layer neural network, each with 200 and 100 nodes. The propensity network $\pi$ has 2 layers each with 200 nodes. The regressor network $\mu$ has 6 layers with 200 nodes and 100 nodes in the first and last 3 layers. All the layers of the VAE and GAN use ReLU activation and the Adam optimizer. The batch sizes are the same as for the adversarial module. We set the learning rate of all the networks as 1e-4 and the hyperparameters $\alpha$ and $\beta$ are set at 1 for all 3 datasets.

\begin{figure*}[t]
\centering
\includegraphics[width=0.8\textwidth]{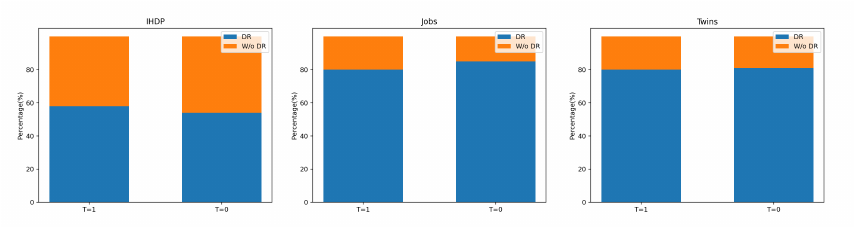} 
\caption{Performance comparison of doubly robust vs. non-doubly robust version of DR-VIDAL. 
% The bar plots show how many times one model setup is better than the other in terms of error on the factual outcome ($y_f$). 
Panels, from left to right, show results on IHDP, Jobs and Twins datasets (100, 10, 100 iterations), respectively.}
\label{fig: boxplots_yf}
\end{figure*}

\paragraph{Implementation and availability.} DR-VIDAL is written in Pytorch (\url{https://pytorch.org/}) and is available under the MIT license at: \url{https://bitbucket.org/goingdeep2406/DR-VIDAL/src/master/}.

\subsection*{A10 Performance of all the various DR-VIDAL configurations} \label{DR_vidal}
The performance of all the various DR-VIDAL configurations are mentioned in Table \ref{tab: big_table}.

\subsection*{A11 DR-VIDAL's performance on IHDP and Twins datasets for $\sqrt{\epsilon_{PEHE}}$ values} \label{IHDP_Twins_comparison}
The performance of various models for $\sqrt{\epsilon_{PEHE}}$ values on the IHDP and Twins dataset are shown in Table \ref{tab: IHDP_Twins_within}.

\begin{table}[h]
\fontsize{7.6pt}{0.35cm}\selectfont
% \small
\centering
\begin{tabular}{l|l r|l r}
\toprule 
    % & \multicolumn{2}{c}{\textbf{IHDP}} & \multicolumn{2}{|c}{\textbf{Jobs}} \\
      & \multicolumn{2}{c}{\textbf{IHDP($\boldsymbol{\epsilon_{ATE}}$)}} & \multicolumn{2}{|c}{\textbf{Twins($\boldsymbol{\epsilon_{ATE}}$)}} \\
      $\boldsymbol{Methods}$ & Out-sample & In-Sample
      & Out-sample & In-Sample\\
\midrule 
      OLS/LR1   & 0.94 $\pm$ 0.06 & 0.73 $\pm$ 0.04  & 0.0069 $\pm$ 0.0056 & 0.0038 $\pm$ 0.0025 \\
      OLS/LR2   & 0.31 $\pm$ 0.02 & 0.14 $\pm$ 0.01  & 0.0070 $\pm$ 0.0025 & 0.0039 $\pm$ 0.0025 \\
      BLR       & 0.93 $\pm$ 0.05 & 0.72 $\pm$ 0.04  & 0.0334 $\pm$ 0.0092 & 0.0057 $\pm$ 0.0036 \\
      k-NN      & 0.90 $\pm$ 0.05 & 0.14 $\pm$ 0.01  & 0.0051 $\pm$ 0.0039 & 0.0028 $\pm$ 0.0021 \\
\midrule 
      BART      & 0.34 $\pm$ 0.02 & 0.23 $\pm$ 0.01 & 0.1265 $\pm$ 0.0234 & 0.1206 $\pm$ 0.0236 \\
      R Forest  & 0.96 $\pm$ 0.06 & 0.73 $\pm$ 0.05 & 0.0080 $\pm$ 0.0051 & 0.0049 $\pm$ 0.0034 \\
      C Forest  & 0.40 $\pm$ 0.03 & 0.18 $\pm$ 0.01 & 0.0335 $\pm$ 0.0083 & 0.0286 $\pm$ 0.0035 \\
\midrule 
      BNN          & 0.42 $\pm$ 0.03 & 0.37 $\pm$ 0.03     & 0.0203 $\pm$ 0.0071 & 0.0056 $\pm$ 0.0032 \\
      TARNET       & 0.28 $\pm$ 0.01 & 0.26 $\pm$ 0.01   & 0.0151 $\pm$ 0.0018 & 0.0108 $\pm$ 0.0017 \\
      CFR$_{WASS}$ & 0.27 $\pm$ 0.01 & 0.25 $\pm$ 0.01    & 0.0284 $\pm$ 0.0032 & 0.0112 $\pm$ 0.0016 \\
\midrule 
      GANITE   & 0.49 $\pm$ 0.05 & 0.43 $\pm$ 0.05 & 0.0089 $\pm$ 0.0075 & 0.0058 $\pm$ 0.0017 \\
      CEVAE    & 0.46 $\pm$ 0.02 & 0.34 $\pm$ 0.01 & n.r & n.r \\
\midrule
      \textbf{DR-VIDAL}  & \textbf{0.69 $\pm$ 0.06} & \textbf{0.57 $\pm$ 0.07} 
      & \textbf{0.0111 $\pm$ 0.0137} & \textbf{0.0102 $\pm$ 0.0128} \\
\bottomrule
\end{tabular}
\caption{Performance of 
various models on the IHDP and Twins datasets for $\epsilon_{ATE}$ (mean $\pm$ st.dev).}\smallskip
\label{tab: IHDP_Twins_within}
\end{table}

\begin{figure}[t]
\centering
\includegraphics[width=0.4\textwidth]{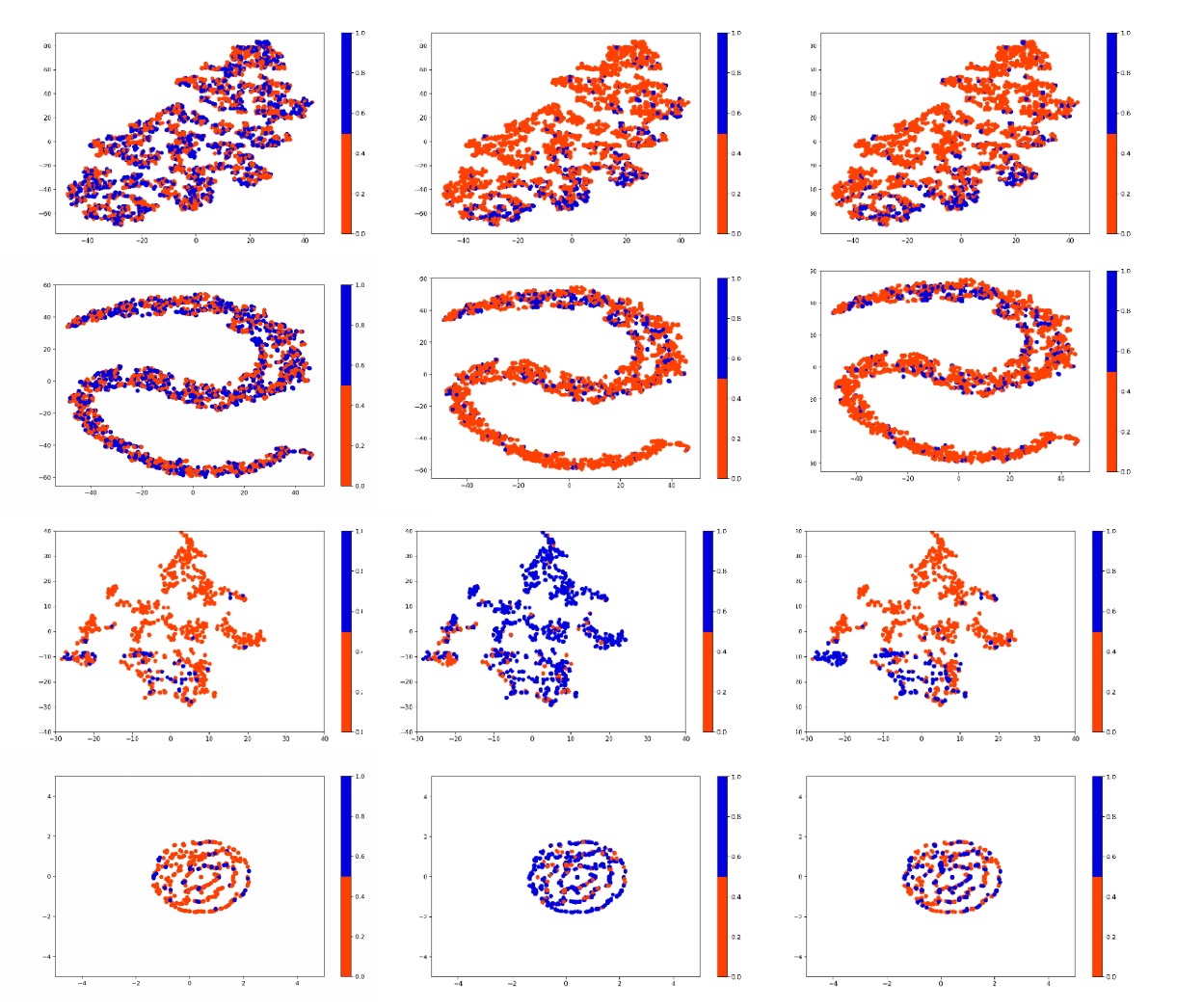} 
\caption{Visualization of the latent representation learned by the VAE module of DR-VIDAL for the Twins and Jobs dataset using t-SNE. The 1$^{st}$ and 2$^{nd}$ panels show the t-SNE before and after training the network for Twins dataset. The 3$^{rd}$ and 4$^{th}$ panels show the same for Jobs dataset. From left to right, the plots show the t-SNE of treatment, factual and counterfactual outcomes.}
\label{fig: Twins_Jobs_latent}
\end{figure}

\subsection*{A10 DR-VIDAL's in-sample performance on Jobs dataset for $R_{Pol}$ values} \label{Jobs_comparison}
The performance of various models on the Jobs dataset for $R_{Pol}$ values are shown in Table \ref{tab: Jobs_within}.

\subsection*{A12 Performance comparison of doubly robust vs. non-doubly robust version of DR-VIDAL} \label{bar_comparison} 
Performance comparison of doubly robust vs. non-doubly robust version of DR-VIDAL is shown in Figure \ref{fig: boxplots_yf}.The bar plots show how many times one model setup is better than the other in terms of error on the factual outcome (yf).

\subsection*{A13 t\_SNE of representations} \label{t_SNE}
The t-distributed stochastic neighbor embedding (t-SNE) of representations learned by the VAE of the adversarial module of DR-VIDAL for Twins and Jobs datasets --before and after training-- are shown in Figure \ref{fig: Twins_Jobs_latent}. For all datasets, the t-SNE shows reorganization and cluster tightness (i.e., the data reside on a smaller space) on the treatment, factual and counterfactual outcomes spaces.

\end{document}